\documentclass[twoside,11pt]{article}
\usepackage{jair, theapa, rawfonts}
\usepackage{enumitem}
\usepackage{color}
\usepackage{algorithm2e}
\usepackage{graphicx}
\usepackage{amsmath}
\usepackage{amssymb}
\usepackage{amsthm}
\usepackage{geometry}
\usepackage{caption}
\usepackage{subcaption}
\usepackage{pbox}
\usepackage{multirow}
\usepackage{array}
\usepackage{xcolor}
\usepackage{soul}

\newcolumntype{L}[1]{>{\raggedright\let\newline\\\arraybackslash\hspace{0pt}}m{#1}}
\newcolumntype{C}[1]{>{\centering\let\newline\\\arraybackslash\hspace{0pt}}m{#1}}
\newcolumntype{R}[1]{>{\raggedleft\let\newline\\\arraybackslash\hspace{0pt}}m{#1}}

\ShortHeadings{Detecting Content-dense News Texts}
{Yinfei Yang, \& Ani Nenkova}
\firstpageno{1}

\begin{document}

\title{Combining Lexical and Syntactic Features for Detecting Content-dense Texts in News\thanks{In submission to JAIR}}

\author{
       \name Yinfei Yang \email yangyin7@gmail.com \\
       \addr Redfin Inc.\\
       2025 1st Avenue\\
       Seattle, WA  98121 USA
       \AND
       \name Ani Nenkova \email nenkova@seas.upenn.edu \\
       \addr University of Pennsylvania \\
       3330 Walnut Street \\
       Philadelphia, PA, 19103 USA
}


\maketitle

\begin{abstract}
Content-dense news report important factual information about an event in direct, succinct manner. Information seeking applications such as information extraction, question answering and summarization normally assume all text they deal with is content-dense. Here we empirically test this assumption on news articles from the business, U.S. international relations, sports and science journalism domains. Our findings clearly indicate that about half of the news texts in our study are in fact not content-dense and motivate the development of a supervised content-density detector. We heuristically label a large training corpus for the task and train a two-layer classifying model based on lexical and unlexicalized syntactic features. On manually annotated data, we compare the performance of domain-specific classifiers, trained on data only from a given news domain and a general classifier in which data from all four domains is pooled together. Our annotation and prediction experiments demonstrate that the concept of content density varies depending on the domain and that naive annotators provide judgement biased toward the stereotypical domain label. Domain-specific classifiers are more accurate for domains in which content-dense texts are typically fewer. Domain independent classifiers reproduce better naive crowdsourced judgements. Classification prediction is high across all conditions, around 80\%\footnote{We reported our initial work on detection of information-dense text in a paper published at AAAI 2014 \cite{yang2014} In this manuscript we have further extended the work by using much larger samples of New York Times articles for training and by analyzing the performance of a larger number of two-layer classifiers. Here we also introduce a new collection of manually annotated test data, for both domain-dependent and general content-density of texts. We make use of this collection for detailed evaluation of the content-density classifiers. With the publication of this manuscript, we will make the classifiers and data available for use by others In our AAAI 2014 paper, we use the term information-dense to describe the types of text we wish to detect. Here we switch the terminology to content-dense, to avoid confusion with work in the intersection of cognitive science and computational linguistics that uses the term information density in information-theoretic sense to describe the change in surprise in the linear processing of sentences \cite{jaeger2010redundancy,pate2015talkers}.}. 
\end{abstract}

\section{Introduction}
\label{Introduction}

News articles are written with different goals in mind. Some aim to inform the reader about an important event, focusing on specific details such as who did what to whom where and when. Others aim to provide background information, facts related to an event and necessary to understand an event but not newsworthy by themselves. Yet others seek to entertain the reader, or to showcase the brilliant mastery of language and the wit of the author. 

In this paper we introduce the task of detecting if a text is content-dense or not. Content-dense news report important factual information about an event, in direct and succinct manner. Prototypical examples of content-dense texts are newswire articles, which are usually perfect answers to a $``$What happened?$"$ question, grounded in a specific event. In general news, however, newswire-like, content-dense text is not the norm.  

We base our analysis on the opening paragraph, called the \textit{lead or lede}, of news articles drawn from the New York Times. News reports often adhere to the inverted pyramid structure, in which the lead conveys what happened, when and where, followed by more details in the body. Information that is not essential is included in the final tail. When writers adhere to this style of writing, the leads are informative and provide positive examples of content-dense texts. Alternatively, the lead may be creative, provocative or entertaining rather than informative, providing examples of non content-dense texts. 

Consider the leads below, from the politics and sports section of the New York Times. The first two are content-dense leads. The other two are non content-dense leads that do not focus on events; they contain few key facts and which are much richer stylistically. 

\begin{normalsize}
\vspace{2mm}	
{
\textbf{Content-dense:}

\textit{\textbf{[Politics]}
Evo Morales, a candidate for president who has pledged to reverse a campaign financed by the United States to wipe out coca growing, scored a decisive victory in general elections in Bolivia on Sunday.
}

\textit{
Mr. Morales, 46, an Aymara Indian and former coca farmer who also promises to roll back American-prescribed economic changes, had garnered up to 51 percent of the vote, according to televised quick-count polls, which tally a sample of votes at polling places and are considered highly accurate.
}

\vspace{2mm}
\textit{\textbf{[Sports]}
North Carolina (29-1) and Duke (26-3) of the Atlantic Coast Conference received No. 1 seedings yesterday in the 64-team women's N.C.A.A. tournament, along with Ohio State (28-2) and Louisiana State (27-3).
}

\textit{
The top-ranked Tar Heels received the No. 1 overall seeding, but were placed in what appears to be the most difficult regional.}

}
\end{normalsize}

\begin{normalsize}
\vspace{2mm}
{
\textbf{Non content-dense:}

\textit{\textbf{[Politics]}
When the definitive history of the Iraq war is written, future historians will surely want to ask Saddam Hussein and George W. Bush each one big question. To Saddam, the question would be: What were you thinking? If you had no weapons of mass destruction, why did you keep acting as though you did? For Mr. Bush, the question would be: What were you thinking? If you bet your whole presidency on succeeding in Iraq, why did you let Donald Rumsfeld run the war with just enough troops to lose? Why didn't you establish security inside Iraq and along its borders? How could you ever have thought this would be easy?
}

\vspace{-1mm}
\textit{
The answer to these questions can be found in what was America's greatest intelligence failure in Iraq -- and that was not about W.M.D.
}

\vspace{2mm}
\textit{\textbf{[Sports]}
With his silver pants and dark blue jersey covered by a mottled mix of grass stains, paint and mud, New England Patriots running back Corey Dillon sat on an aluminum bench on the sideline at Gillette Field on Sunday, looking exhausted and frozen.
}

\vspace{-1mm}
\textit{
Only a few minutes remained in the Patriots' 20-3 victory over the Indianapolis Colts, and Dillon was resting. He stared at the field, snowflakes swirling around his head as the realization of his first playoff victory swirled inside it.
}

}
\end{normalsize}

\vspace{2mm}

Below we propose an approach for labeling short news texts as content-dense or not. Our analysis of manual annotations reveals that uninformative article leads are common. We investigate several types of lexical and non-lexicalized syntactic features for distinguishing content-dense texts from other more general or creatively written texts. We present a two-layer ensamble classifier model which significantly outperforms a baseline assuming that all news leads are content-dense. We also study the robustness of the definition of content density across domains, as well as the performance of domain-dependent and domain-independent (general) classifiers.

\section{Corpus}
\label{data}

The data for our experiments comes from the New York Times (NYT) annotated corpus (LDC Catalog No. LDC2008T19). The corpus contains 20 years worth of NYT editions, along with rich meta-data about the newspaper section in which the article appeared and  summaries produced by information scientists for many of the articles. The leads of articles are explicitly marked in the corpus, so extracting the relevant text for further analysis is straightforward. 

In our previous proof-of-concept work \cite{yang2014}, we selected a subcorpus of articles published in 2005 or 2006 from four different genres (business, U.S. international relations, science and sports). Given the selection criteria, the data in that prior work contained considerably fewer articles from the science and the sports domains compared to the other two domains.  Moreover, the performance of the content-dense classifiers in the science and sports domains was notably worse than the other two domains, which could be explained either by the fact that these classifiers were trained on smaller datasets or by the intrinsic difficult of predicting content density in these two domains. To definitively  resolve this question, and to benefit from the largest training dataset possible, we extend the corpus to the full NYT corpus in the experiments reported in this manuscript. 

We also expect that the degree to which a text would be judged to be content-dense, reporting on important event in a direct manner, is influenced by the domain of the article. It is reasonable to expect that typical events in science or sports would not be considered of the same importance as international political or business events. To study the cross-domain differences, we analyze four news domains: Business, Sports, Science\footnote{The science articles are from the CATS corpus \cite{Annie2013}, which only contains articles published after 1999.} and US International Relations (or Politics for short).

\subsection{Training set heuristic}

To automatically label leads as content-dense or not, we make use of the manual summaries which accompany many articles in the NYT corpus. For the articles with content-dense leads, the manual summary will be very similar to the lead itself, as this type of lead by definition provides a fact-focused summary of the article. For leads that simply seek to engage the reader via more creative devices, the manual summary will differ considerably from the lead. Overall, the similarity between the lead and the manual summary provides a strong indication of the importance and factual, event-oriented, nature of the information expressed in the lead.  

For articles with manual summaries of at least 25 words, we calculate an content-dense score. For each word in the summary, a tuple $t(w, pos)$ is created containing the word and its part of speech. The score is computed as: 

\begin{align}
    Score = \frac{ \text{ \# of t(w, pos) also in leads} }{ \text{\# of t(w, pos)} }
\end{align}

\subsection{Label analysis}

Table \ref{tab:nyt_detail} shows details about the number of all NYT articles from each of the four domains. The first column shows the number of articles in the NYT from the given domain. The second column shows the number of articles used for training domain-dependent classifiers (we explain the selection procedure below). Overall only about one third of articles have associated manual summaries.

\begin{table}[!htb]
    \centering
    \caption{Number of articles in the corpus.}
    \label{tab:nyt_detail}
    \begin{tabular}{| c | c | c |}
    \hline
    & \textit{Total number of articles} & \textit{Articles used in training (Percentage)} \\ \hline
    Business & 149,113  & 21,224 (14.2\%) \\ \hline
    Science & 23,240    & 7,737 (33.\%)   \\ \hline
    Sports & 134,925    & 10,670 (7.9\%)  \\ \hline
    Politics & 45,926   & 10,503 (22.8\%) \\ \hline 
    \hline
    Overall & 353,204   & 50,134 (14.2\%) \\ \hline
    \end{tabular}
\end{table}

\begin{figure}[!h]
    \centering
    \begin{subfigure}[b]{0.49\textwidth}
            \centering
            \includegraphics[width=\textwidth]{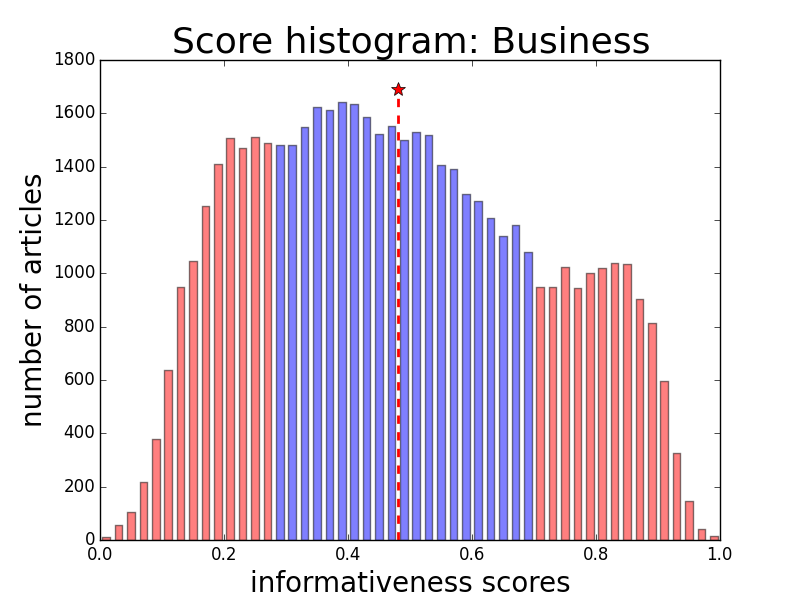}
    \end{subfigure}
    \begin{subfigure}[b]{0.49\textwidth}
            \centering
            \includegraphics[width=\textwidth]{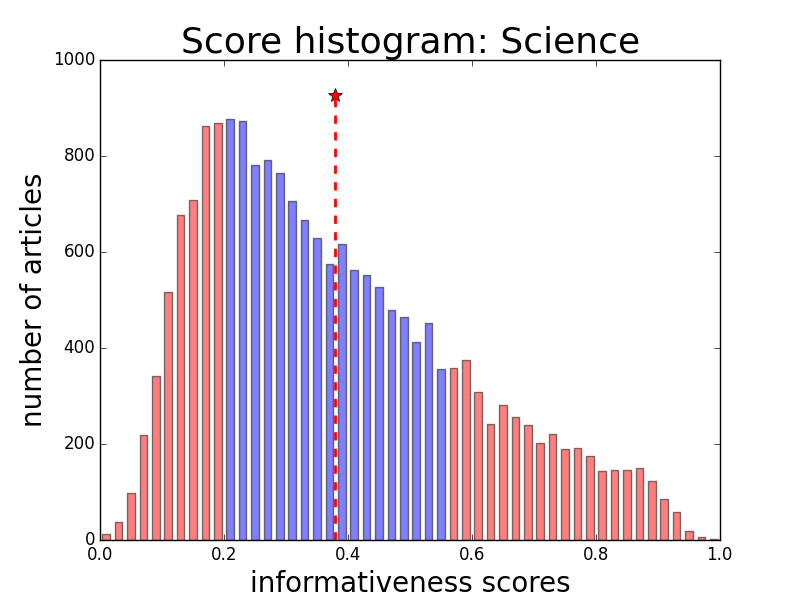}
    \end{subfigure}
    \begin{subfigure}[b]{0.49\textwidth}
            \centering
            \includegraphics[width=\textwidth]{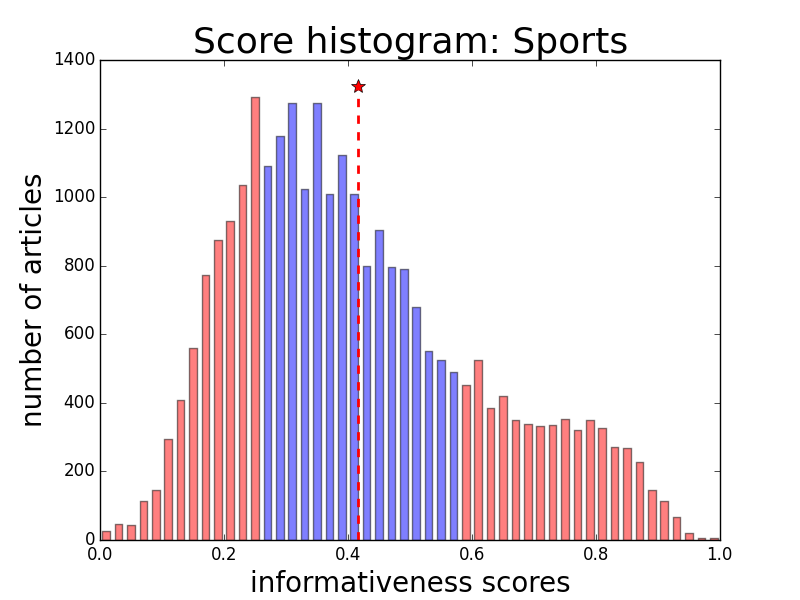}
    \end{subfigure}
    \begin{subfigure}[b]{0.49\textwidth}
            \centering
            \includegraphics[width=\textwidth]{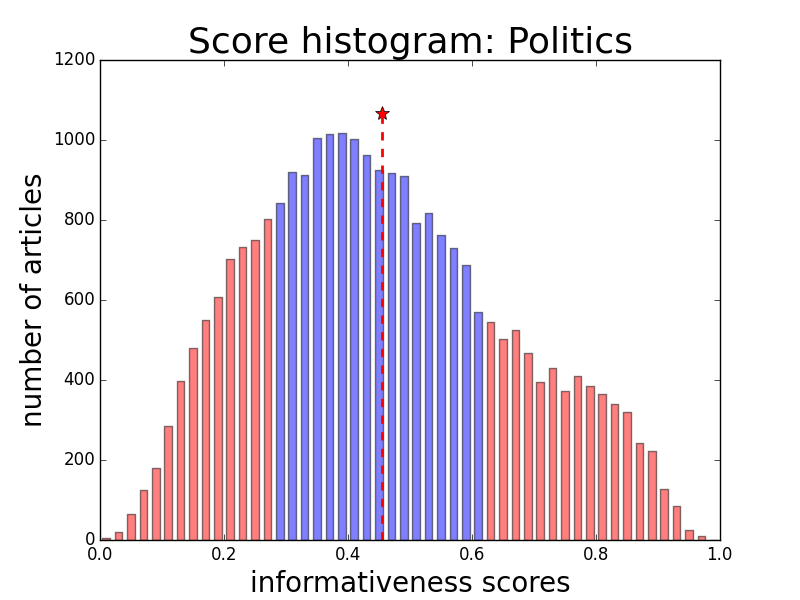}
    \end{subfigure}
    \caption{Score histograms for the four genres: \textbf{[Top Left]} Business, \textbf{[Top Right]} Science, \textbf{[Bottom Left]} Sports, \textbf{[Bottom Right]} Politics. 20th and 80th percentiles are colored red. Red star indicates the average content-dense score for each genre.}
    \label{fig:hist_score}
\end{figure}

The distribution of content-dense scores assigned as a function of the overlap with the human summaries is shown in Figure \ref{fig:hist_score}.
In the business domain the distribution of scores is almost uniform, reflecting the fact that in that section there are articles about important events---company mergers, unexpected stock price changes, product announcements and lawsuits---but also non-event specific analysis of current trends, minor events such as auctions and people-centered pieces about prominent business men and women. 

In sports and science, the distribution of content-dense scores is clearly skewed towards the non content-dense end of the spectrum. In these domains there are fewer intrinsically important events to begin with and writers more often resort to the use of creative and indirect language meant to provoke readers' interest.

The content-dense scores in politics is almost normally distributed, with mean roughly in the middle of the possible range, and much higher than any of the other domains. The non content-dense leads in this domain usually provide a commentary on an ongoing event rather than reports of a specific new development.

In the rest of the paper, we focus on the binary classification task of predicting if a lead is content-dense or not. However, it is reasonable to expect that our indirect labeling scores are noisy. To obtain cleaner data for training our model, we label only the leads with most extreme scores: we assign the label non content-dense to the leads with scores that fall below the 20th percentile and label content-dense to leads that score above the 80th percentile for their domain. The 20th/80th percentile sets are colored red in figure \ref{fig:hist_score}. In the general (domain-independent) model, the data is pooled together and again the leads with lowest scores are assigned to the non content-dense class and the leads with highest scores are considered content-dense.

\section{Methodology}
\label{method}

In this section, we introduce the features and models we used in our experiments. In our prior experiments \cite{yang2014}, we found that lexical features are well-suited for the task, particularly lexical representations determined independently of the training data. Along with these, unlexicalized syntactic representations also lead to remarkably good results. A number of other representations we experimented with did not appear to be that beneficial for the task. Motivated by these findings, here we study in depth the lexical representations and the unlexicalized syntactic representation, and explore ways to combine the predictions of these models to achieve even better accuracy.

\subsection{Features}

We compare and combine two lexical and one syntactic representation. For the lexical representation, we use the  vocabulary from the  \textbf{MRC Database(MRC)}, which is independent of our training set and a vocabulary derived from the training set and weighted by \textbf{Mutual Information(MI)}. The syntactic representation is simply the list of  \textbf{Production Rules(PR)} from the constituency parse of the sentences in the lead.

\paragraph{MRC Database(MRC):} The MRC Psycholinguistic Database \cite{MRC:88} is an electronic dictionary containing 150,837 words, different subsets of which are annotated for 26 linguistic and psycholinguistic attributes. We select a subset of 4,923 words normed for age of acquisition, imagery, concreteness, familiarity and ambiguity. In \cite{MRC:88}, the words were chosen among those with medium frequency in a large corpus and experiment subjects were asked to rate on a scale the degree to which each word has one of these properties. The MRC dictionary is a compilation of results from different studies, run by different research groups, with different criteria for selecting the list of words for which to solicit norms. We use the list of words which have at least one of above ratings. 
The value of each feature is equal to the number of times it appeared in the lead, divided by the number of words in the lead.

About 90\% of the MRC vocabulary (4,647 words) appears at least once in the training data. About 4,300 appear more than five times.



\paragraph{Mutual Information(MI):} The lexical representation described above is domain independent, determined without any knowledge about the data which will be used for training and testing of our classification models. We also introduce a domain-dependent lexical representation, derived from the training data for the classifiers and using  mutual information to measure the association between particular words and the content-dense and non content-dense writing styles. For each genre, we compute the mutual information between words and lead type in the training data as:

\begin{align}
    &\text{MI}_{\text{c}} = \text{log}\frac{p(\text{word}, \text{c})}{p(\text{word})p(\text{c})}
\end{align}

Here $c$ is either the content-dense or the non-content dense class. We only compute the MI scores for words that appear at least  5 times in the training set. We select the top 500 words with highest associations with each of the writing styles, for a total of 1,000 features. The value of the feature is 1 if the word occurs in the lead and 0 otherwise.


The words with highest mutual information\footnote{We ran 10 fold cross validation in the experiments. The mutual information is computed separately based on the training set of each fold. The words listed in table \ref{tab:mi_examples} are from fold 0, but high mutual information words from other folds are very similar.}  with the content-dense classes and non content-dense classes are listed in Table~\ref{tab:mi_examples}. 

The words with high mutual information with the content-dense class are distinctly domain specific. Content-dense leads in the business domain are more likely to talk about companies and their executives, deals, agreements and offers. Content-dense leads in science are more likely to discuss a specific study or drug, since they are overwhelming biased towards health-related topics. Sports content-dense leads are associated with specific sport events  or deals. In politics, content-dense leads discuss American involvement and attacks.

In addition, the words $``$yesterday$"$ and $``$today$"$ also appear among those associated with content-dense leads, providing a strong indicator that the news is focused on a specific recent event rather than a general discussion or personal aspects story. The words associated with the non content-dense class in contrast tend to be related to non-specific activities ({\it find, feel, hear, smile, remember, sit, wait}) and focused on personal aspects rather than on the professional roles({\it  man, people, friend, husband, guy, kid, child, friend}).

Only about half of the words in the mutual information representation also appear in the MRC. 



\begin{table}[!htb]
    \centering
    \caption{Top 30 selected words for each domain and overall data}
    \begin{tabular}{| c | C{6.5cm} | C{6.5cm} |}
    \hline
    & \textit{Content-dense} & \textit{Non content-dense} \\ \hline
    Business & \small\textit{company, yesterday, million, billion, today, percent, group, announce, executive, plan, share, corporation, york, part, deal, agree, largest, unit, court, agency, inc., commission, bank, include, firm, chief, agreement, chairman, offer, service}
    		  & \small\textit{day, stock, ago, work, thing, good, investor, year, find, turn, long, man, economy, job, people, home, street, room, time, rate, lot, index, city, sit, mr., market, wall, money, ms., life}
		  \\ \hline
    Science & \small\textit{study, health, today, yesterday, report, drug, official, research, federal, state, scientist, administration, disease, researcher, company, government, accord, human, virus, university, group, million, expert, announce, cell, include, cancer, united, agency, issue}
    		  & \small\textit{day, mr., ms., ago, feel, hear, room, sit, walk, home, eye, life, friend, thing, run, talk, live, game, stand, back, family, hand, foot, good, morning, husband, hour, night, town, son}
		  \\ \hline
    Sports & \small\textit{yesterday, today, league, team, national, million, season, association, official, die, cup, contract, race, year, deal, tonight, game, conference, major, president, round, lead, charge, announce, committee, victory, win, woman, world, series}
    		  & \small\textit{fan, ago, watch, back, stand, ball, day, question, good, turn, moment, room, smile, feel, hand, time, wear, people, knicks, hear, remember, n.b.a., net, guy, sit, thing, stadium, shot, kid, walk}
		  \\ \hline	
    Politics & \small\textit{official, united, today, american, states, administration, mr., clinton, military, government, weapon, international, effort, attack, security, nuclear, force, report, intelligence, group, court, defense, nations, program, include, china, agency, secretary, nato, plan}
    		  & \small\textit{man, day, world, war, people, time, u.s., ago, back, sit, thing, front, live, city, child, street, room, stand, saddam, morning, america, word, year, wait, car, kerry, young, friend, watch, hour}
		  \\ \hline
    \hline
    Overall & \small\textit{yesterday, today, company, million, official, billion, group, united, percent, announce, states, plan, administration, york, include, american, agency, government, federal, report, accord, court, executive, national, drug, part, state, international, corporation, deal}
    		  & \small\textit{day, ago, thing, man, good, stock, time, room, sit, back, stand, turn, watch, street, hear, home, feel, people, long, life, lot, ms., walk, town, wall, word, friend, live, moment, eye}
    \\ \hline
    \end{tabular}
    \label{tab:mi_examples}
\end{table}

\paragraph{Production Rules(PR):} Finally, we use production rules as the syntactic representation \cite{Annie:12,ganjigunteashok-feng-choi:2013:EMNLP,P13-2150,malmasi2014chinese}.

We view each sentence as the set of grammatical productions, $\textit{LHS} \rightarrow  \textit{RHS}$, which appear in the syntactic parse tree of the sentence. We keep only non-terminal nodes, excluding all lexical information, so the lexical and syntactic representations capture non-overlapping aspects of writing style. All production rules from the training set are used in the representation. The numbers of production rules vary for the four domains, from 16,000 rules (Science) to 32,000 rules (Business).\footnote{Stanford CoreNLP package~\cite{corenlp} is used to extract production rules.}


\subsection{Classifier combination}

The three feature representations we introduced capture domain independent lexical clues for content-density, domain-dependent indicators for important events and general style of writing captured by the structure of sentences in the text. We train a support vector logistic regression classifier with each class of features individually. Furthermore in this section, we examine two approaches for combining the predictions from the three classes of features.

\begin{figure}[!h]
    \centering
    \begin{subfigure}[b]{0.49\textwidth}
            \centering
            \includegraphics[width=\textwidth]{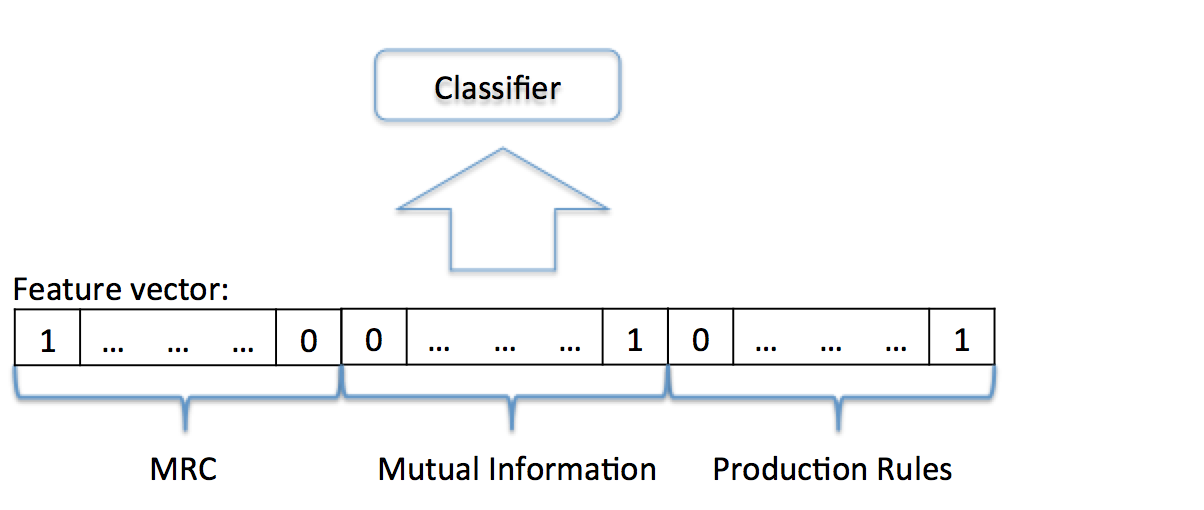}
            \caption{Feature-level combination}
    \end{subfigure}
    \begin{subfigure}[b]{0.49\textwidth}
            \centering
            \includegraphics[width=\textwidth]{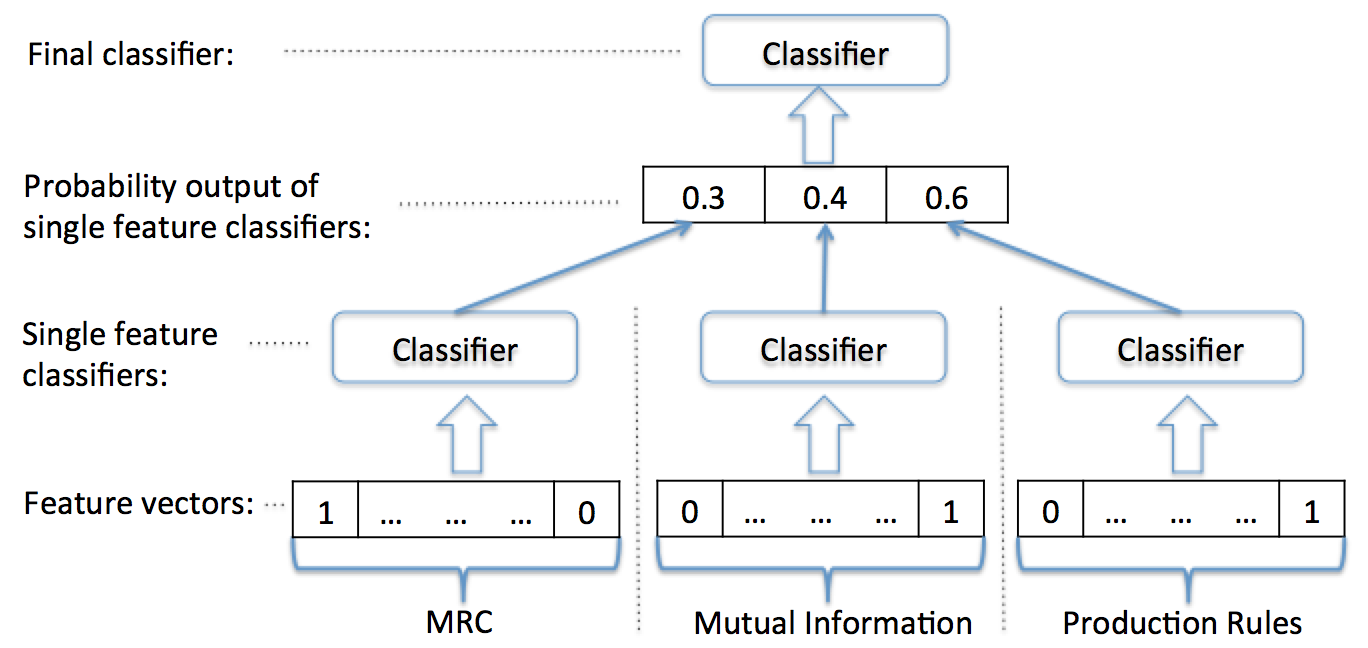}
            \caption{Decision-level combination}
    \end{subfigure}
    \caption{Illustration of feature-level combination and decision-level combination}
    \label{fig:fusions}
\end{figure}

\paragraph{Feature-level combination}
First we examine the performance of feature-level combination to develop a system that makes use of all three types of indicators of content density. We concatenate the three feature representations together in a feature vector. The number of entries in the feature vector is equal to the sum of the number of features of the MRC, mutual information and production rule representations. 
Then we train a logistic regression model  based on the concatenated feature representation. This way of combining evidence lead to overall improvements in our early work. However much work on ensemble learning has demonstrated that  for variety of tasks this method of combination is not as powerful as decision-level combination (For example see  \cite{Raaijmakers:2008:MSA:1613715.1613774,vanHalteren:1998:IDD:980845.980928,DBLP:conf/icassp/MetallinouLN10,1699337}). We treat the feature-level combination as the baseline for our experiments. Figure~\ref{fig:fusions} (a) shows the structure of feature-level combination classifier.

\paragraph{Decision-level combination}

Classifier combination has been shown to outperform feature combination in a single classifier \cite{tulyakov2008}. There are multiple reasons why this may be the case, especially for a linear classifier like the one we use. Concatenating all features in a single representation makes the system prone to over-fitting, as the number of features becomes closer to the number of training examples. If the number of features of a given type is considerably smaller (for example there are many more features in the production rule representation compared to the mutual information representation), the signal contributing to the final decision may be dominated by the larger class, defeating the purpose of evidence combination. It could also lead to the presence of correlated features, for example in the combination of the two types of lexical features.

We propose a two layer classifier combination system. We first train a logistic regression classifier with each of the three feature representations individually. Then another model is trained, in which the features are the probabilities of the content-dense class from the first layer classifiers. In the experiment, the corpus is split into training set, development set and testing set. The first layer classifiers is trained on the training set, and the second layer classifier is trained on development set. Figure~\ref{fig:fusions} (b) illustrates the structure of the decision-level combination system.

\section{Evaluation on automatic annotations}

\subsection{Classifier evaluation}
In the feature-level combination system, we train the binary classifier using Liblinear \cite{liblinear} with linear kernel and L2-regularized logistic regression model setting. In the decision-level combination experiments, we first train binary classifiers based on each feature representation using LibLinear with the same settings. 
Using the probability outputs (for the content-dense class) of the first stage classifiers as features, we then train a final binary classifier using LibSVM \cite{libsvm} with linear kernel. Grid search is used to find the best hyper-parameters in all models.

We perform 10-fold cross-validation experiments on the entire heuristically labeled data, where five folds are used for training first-stage classifiers and the feature-level combination classifier.  Four folds are used for training the second-stage combination classifier, which uses only the probabilities of the content-dense class from the first stage classifiers.  One fold is used for testing the classifiers. We evaluate the two combination models on the automatically labeled data but also analyze the performance when only a single class of feature is used. 

The results are presented in Table \ref{tab:res}. Because of the way the data was labeled, the two classes are of equal size, with $50\%$ accuracy as the random baseline.
The top three rows in the table corresponds to a system trained with only one class of features. The last two rows shows the results for the two combination systems. The columns correspond to the domains we study---business, science, sports and politics.
The domain-specific models were trained and tested only on the data from the given domain and the results are shown in the first four columns. The general, domain-independent model is trained and tested on the combined dataset and the last column shows its performance. 

Depending on the domain, accuracies are high, ranging between $87.2\%$ for business and $83.6\%$ for politics.

Of the individual feature classes,  the production rules representation leads to the best overall accuracy. 
Combining the representations at the feature level leads to improvement over the production rule classifier for the business and politics domain, as well as in the general domain-independent classifier but not for science and sports where performance using all features is in fact worse than using production rules alone.

In line with our previous work, all single feature classifiers have very good performances. The production rules (PR) syntactic representations lead to the best performance for all domains, with accuracies over or close to $80\%$ for all domains. The most important rules are quite different in each genre, but the discovered patterns are mostly aligned with our intuition. For example, \textit{VP -\textgreater VB NP PRT ADVP} is often associated with content-dense leads in Business, the example text like \textit{VP -\textgreater VB[push] NP[the Czech currency] PRT[up] ADVP[sharply]}. The rule \textit{NP -\textgreater JJ CD NNS}, however, is usually associated with non content-dense leads, e.g. \textit{NP -\textgreater JJ[pre-April] CD[15] NNS[blues]}. The production rules with highest weights are listed in Appendix B.

Of the lexical representations, the MRC representation leads to better results, with accuracies varying  from $82.7\%$ for the business domain to $79.4\%$ for the sports domain. The corpus-dependent lexical representations based on mutual information has a slightly lower performance: the accuracies range between $81.9\%$ for business and $78.1\%$ for politics. 

The results for the general classifier---which is trained and tested on data from the four domains pooled together---are similar. For this classifier leads may change their labels, for example a sports article whose content-dense score is in the 80th percentile of scores for sports may fall below the 20th percentile when all data is combined. 

The fact that the representations designed independently of the training data can lead to such good results is a positive finding, indicating that the results are likely to be robust. 

For all the domains and general domain-independent data, decision-level combination considerably improves the performance compared to  classifiers trained with only one of the representations. It is the most accurate among the five classifiers that we compare, with up to $3.8\%$ performance gain in politics compared to the best single feature classifier. 

The baseline combination system, feature-level combination, performs worse than the decision-level combination. One of the possible reasons is that given the increased number of features, this model may require more training data to reach its performance potential. We study this aspect of model development in section \ref{sec:training_data}.

\begin{table}[!htb]
    \centering
    \caption{Binary classification accuracy(\%) of 10-fold cross validation on the automatically labeled set for different classes of features and two fusion models}
    \label{tab:res}
    \begin{tabular}{| c | c | c | c | c | c |}
    \hline

    & \textbf{Business} & \textbf{Science} & \textbf{Sports} & \textbf{Politics} & \textbf{General} \\ \hline
    MRC  Database (MRC)           & 82.7 & 81.1 & 79.4 & 78.5 & 82.0 \\ \hline
    Mutual Information (MI)           & 81.9 & 79.8 & 78.8 & 78.1 & 80.6 \\ \hline
    Production Rules (PR)            & 83.8 & 83.8 & 82.6 & 79.8 & 83.5 \\ \hline
    \hline
    Feature-Level Fusion              & 85.5 & 81.4 & 81.5 & 80.8 & 85.9 \\ \hline
    Decision-Level Fusion            & \textbf{87.2} & \textbf{87.0} & \textbf{85.6} & \textbf{83.6} & \textbf{86.7} \\ \hline
    \end{tabular}
\end{table}

\subsection{Combining classifiers with different representations}

Here we evaluate different possible combinations of feature types. We compare these possibilities for decision-level combination, which we already established works better than feature-level combination. 

The motivation to examine combinations of features is that not all features are available in all applications. Moreover concerns about run time may make syntactic features undesirable in certain settings, where syntactic parsing may not be feasible. Mutual information representations also require larger training data for each domain of interest, to compute the mutual weights for each feature. So we examine the effectiveness of combining different feature classes. The multilayer structure makes the decision-level fusion easier to add or remove features.  Developers can simply train a classifier based on new features, then add them to the second layer without affecting existing single feature classifiers. 

We show the results from evaluating three different classifier combinations: MRC+MI (lexical features only), MRC+PR (domain independent features only) and MRC+MI+PR (all features together).

The results are shown in Table \ref{tab:res_comb}. The top row in the table corresponds to the baseline, feature-level combination model with all three classes of features. Rows 2-4 correspond to decision-level models with the three different classifier combinations. As in previous tables, the first four columns correspond to domain-specific models, and the last column shows the results for the general, domain-independent model. Combination classifiers based on all three features in decision-level combination still has the highest accuracy, showing that each of the three representations contributes to the improved performance of the classifier.  The domain independent features, MRC+PR with decision-level combination shows a competitive results too, suggesting that the mutual information representation is the one that could be removed with least degradation in performance. The accuracies are just slightly lower than the best,  $0.5\%$ lower for the business domain for example. 

The decision-level combination of lexical representations has lower performance then the other two decision-level combination models. The accuracies range between $84.5\%$ for the business domain and $80.3\%$ for the politics domain. The combination of the two lexical representation leads to better performance than using either of the individual features classes, suggesting that MRC+MI combination at the decision-level is a good alternative when syntactic features are not available.

\begin{table}[!htb]
    \centering
    \caption{Binary classification accuracy(\%) of 10-fold cross validation on the automatically labeled set for different combinations of features}
    \label{tab:res_comb}
    \begin{tabular}{| c | c | c | c | c | c |}
    \hline

    & \textbf{Business} & \textbf{Science} & \textbf{Sports} & \textbf{Politics} & \textbf{General} \\ \hline
    Feature-Level Fusion            & 85.5 & 81.4 & 81.5 & 80.8 & 85.9 \\ \hline
    \hline
    MRC+MI                   & 84.5 & 82.5 & 81.6 & 80.3 & 83.5 \\ \hline
    MRC+PR                  & 86.7 & 86.2 & 84.7 & 83.0 & 86.1 \\ \hline
    MRC+MI+PR            & 87.2 & 87.0 & 85.6 & 83.6 & 86.7 \\ \hline
    \end{tabular}
\end{table}

\subsection{Is the training data enough?}
\label{sec:training_data}

 We now discuss the impact  of the training set size on classifier performance.  We evaluate the relationship between classifier accuracy and the  increasing of the number of training instances for each domain. We start with a training set of 100 articles, growing to 6,500 instances in the training data, increasing the training set with 100 randomly selected articles in each step.
 Accuracy is computed on the same testing set for each domain. As in our previous experiments, 10-fold cross validation is performed. For each fold, there is a dedicated test set, which means all cross-validation iterations used the same test set. The reported results are an average of the accuracies on the fixed test set in each fold.
 
\begin{figure}[!h]
    \centering
    \begin{subfigure}[b]{0.49\textwidth}
            \centering
            \includegraphics[width=\textwidth]{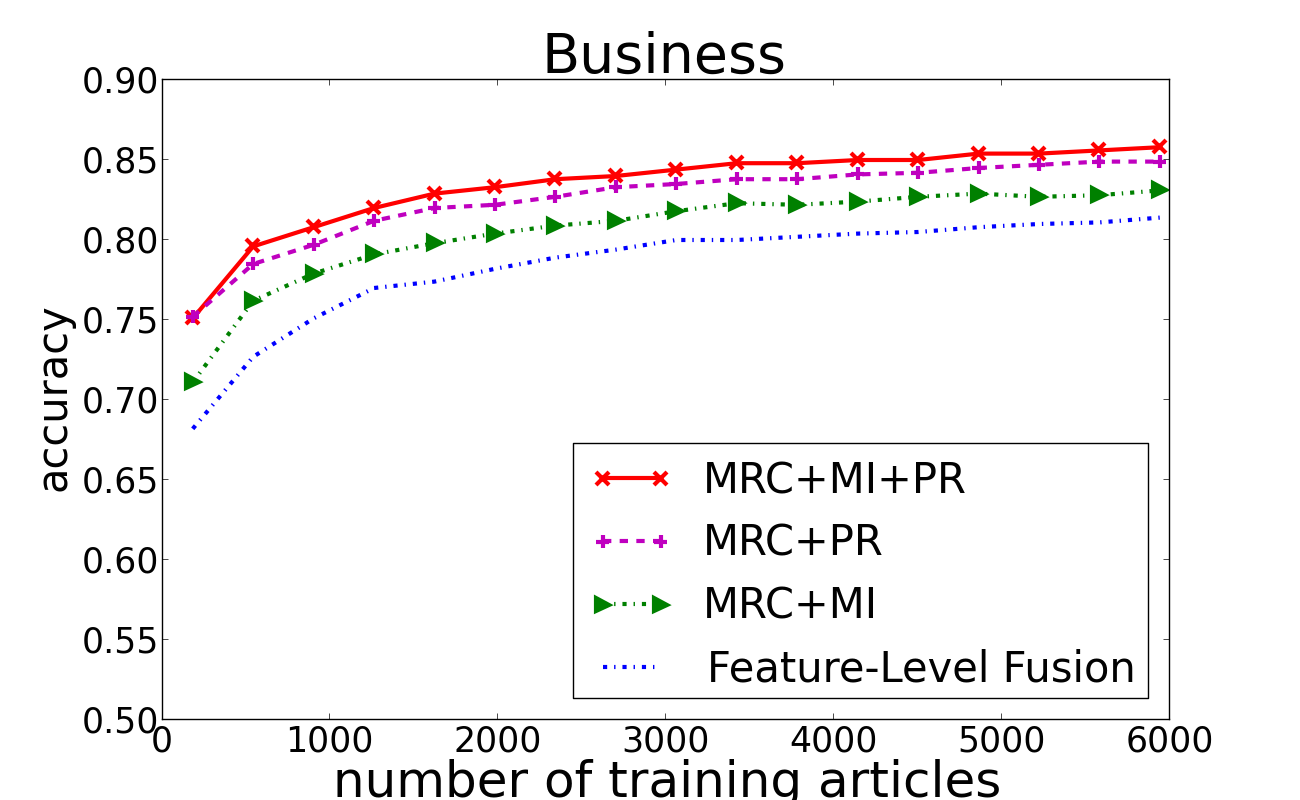}
    \end{subfigure}
    \begin{subfigure}[b]{0.49\textwidth}
            \centering
            \includegraphics[width=\textwidth]{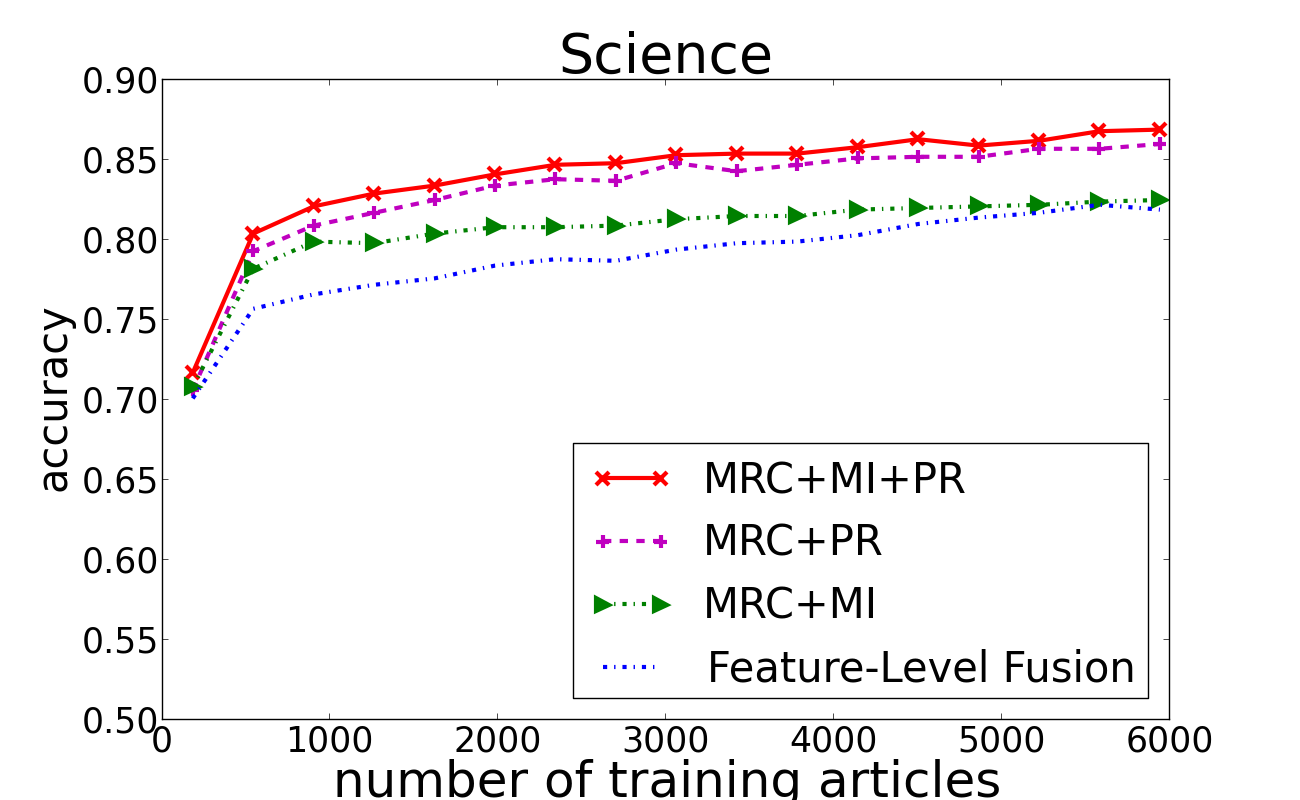}
    \end{subfigure}
    \begin{subfigure}[b]{0.49\textwidth}
            \centering
            \includegraphics[width=\textwidth]{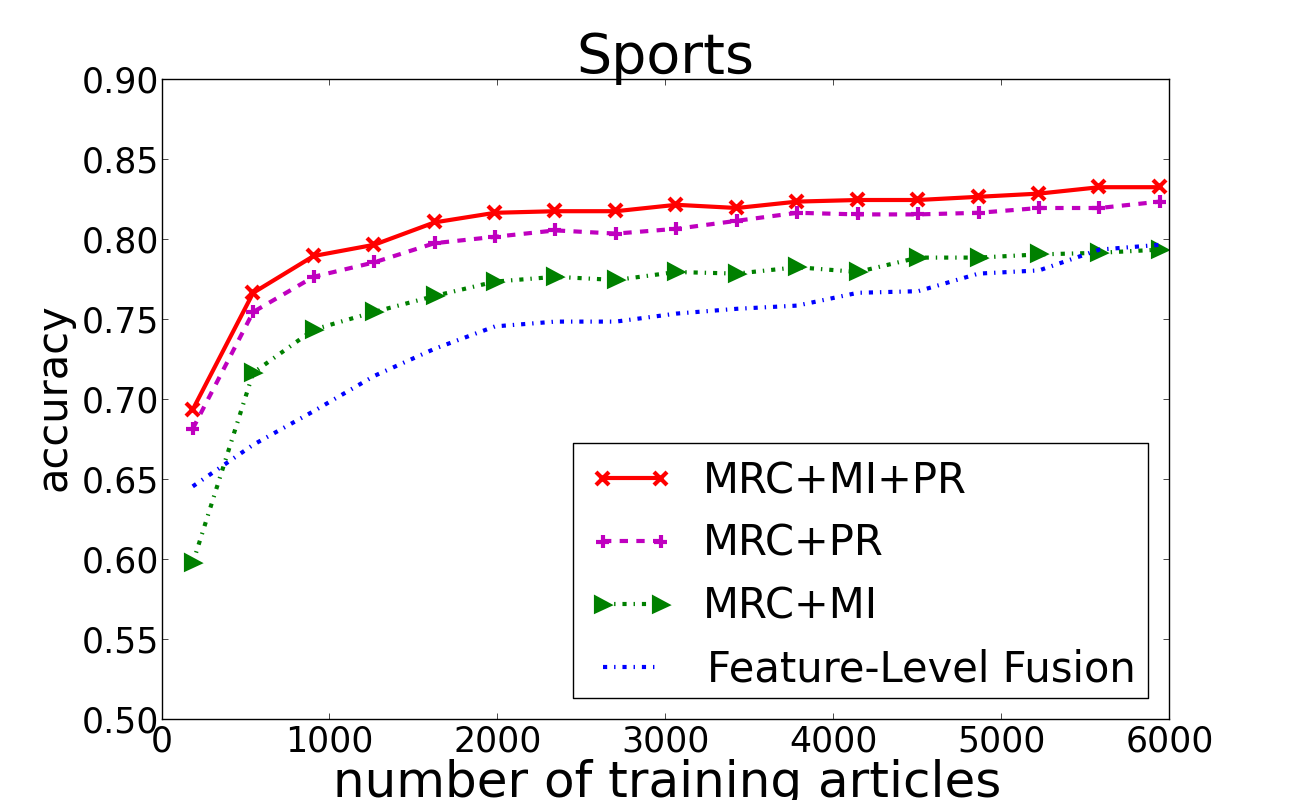}
    \end{subfigure}
    \begin{subfigure}[b]{0.49\textwidth}
            \centering
            \includegraphics[width=\textwidth]{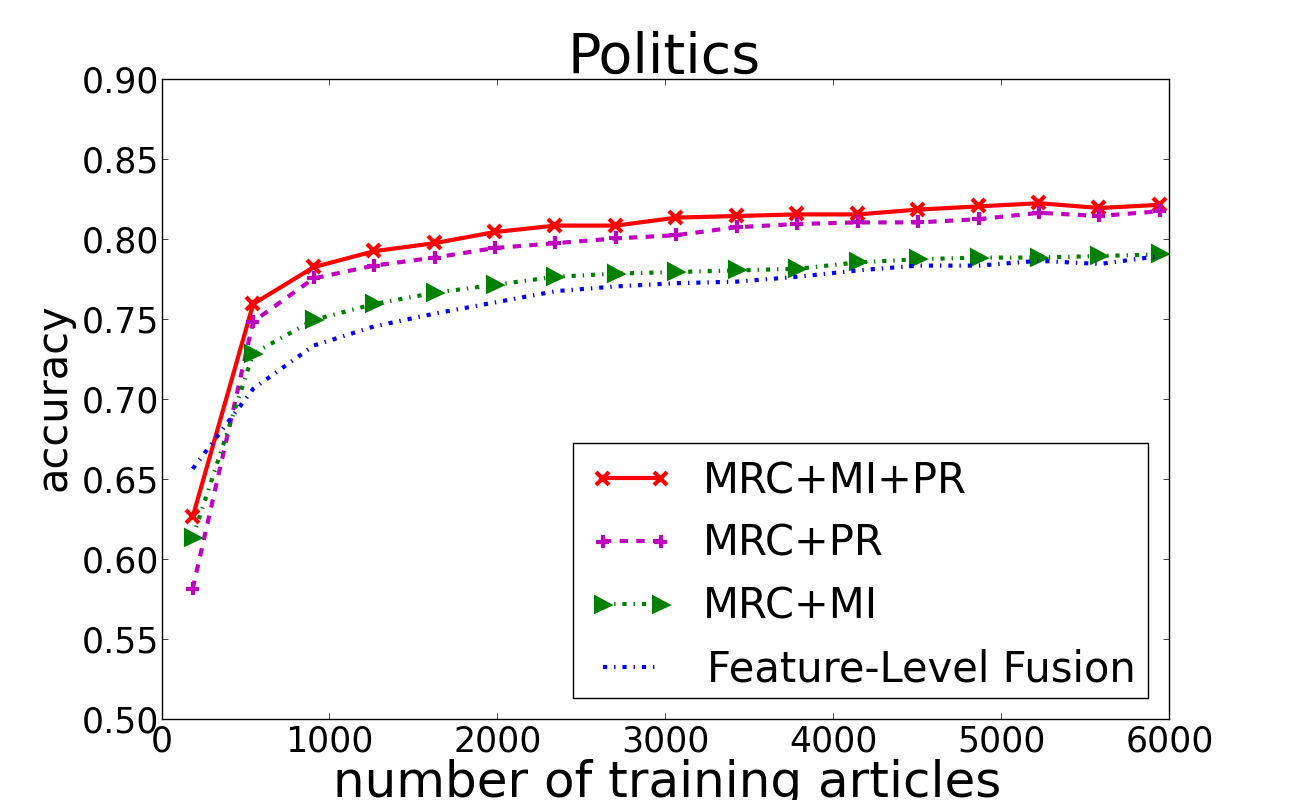}
    \end{subfigure}
    \caption{Accuracy by changing size of training set for the four genres: \textbf{[Top Left]} Business,  \textbf{[Top Right]} Science,  \textbf{[Bottom Left]} Sports,  \textbf{[Bottom Right]} Politics}
    \label{fig:increase_number}
\end{figure}

Figure \ref{fig:increase_number} shows the accuracy/size curve for each domain. Among the four genres, decision-level combination of all three features has the highest accuracy. The accuracy increases rapidly with the increase of training data when the number of training articles is less than 2,000. When the size is larger than 2,000, it continues to increase, but very slowly. The decision-level combination of MRC+PR features, which is the second best model for all domains, behaves similarly. The accuracy of the MRC+MI  decision-level combination is the worst of the combination systems and exhibits the slowest increase. 

The accuracies of decision-level combination with 6,500 training article are already very close to the final numbers with full training set (shown in table \ref{tab:res_comb}). Increasing the number of training instances barely changes the performance after this point. 

The baseline, feature-level combination, has the lowest accuracies. Yet we still see increase in accuracy as the training set size increases. For three of the domains, its performance becomes the same as that of the MRC+MI combination with a large enough training set.

The results also indicate that decision-level combination is able to achieve better performance with less training data. 

The graphs suggest that the difference in performance of the content-density predictor in the four domains likely reflects the difficulty of the domain rather than the difference in training data size.

\section{Evaluation on human annotations}

So far we have  established that recognition of content-dense texts can be done very accurately when the label for the lead is determined  by intuitive heuristics on the available article/summary resources. We would like however to test the models on manually annotated data as well, in order to verify that the predictions indeed conform to reader perception of the style of the article.

We selected a total of 1,000 articles and split them into two sets. For the first set of 400 articles, the authors of the paper annotated the content-dense labels and provided a real-value score for the domain-dependent content-density of each text. Then a second set of 600 articles was selected and annotated on Amazon Mechanic Turk (AMT).
All annotated articles were randomly picked from the NYT data and did not appear in the training data for the classifiers that we evaluate here.

\subsection{Human annotated dataset}
\subsubsection{Basic set}
In the basic human annotation set, the authors of the paper annotated 400 NYT articles, 100 from each domain, with judgements of their perceived informativeness. Similar to prior work on grammatically judgements \cite{bard}, the annotation was done with respect to a reference lead that fell around the middle of the content-dense spectrum. Leads were labeled by domain: the question was if a specific article from domain $D$ is content-dense compared to the reference lead for that domain. All 100 leads from the same domain were grouped together and displayed in random order, with the annotators seeing leads only from the same domain until they completed the annotation for that domain. The reference lead in each case was drawn from the respective domain. 
The annotator gave both a categorical label for the lead (less content-dense or more content-dense than the reference) and a real value score (ranging between 0 to 100) via a sliding bar. The categorical labels were used to test the binary classifiers. The real-valued annotations were used to compute correlations with classification scores produced by the classifier.

\paragraph{Inter-annotator agreement} 
All 400 test leads were annotated as being content-dense or not and with a real-value indicator of the extent to which they are content-dense. 
Table \ref{tab:human_agreement} shows the percent agreement between the two annotators on the binary level task, as well as the correlation of the real-value annotation. For the binary annotations we also report the Kappa statistic.  

\begin{table}[!htb]
    \centering
    \caption{Inter-annotator agreement on manual annotations. Percent agreement is computed on the binary annotation, correlation is computed on the real-value degree of content-density of the leads. All correlations are highly significant, with $p < 0.001$.}
    \label{tab:human_agreement}
    \begin{tabular}{| c | c | c | c | }
    \hline

    & Agreement & Kappa & Correlation \\ \hline
    Business & 0.70 & 0.405 & 0.608 \\ \hline
    Science   & 0.74 & 0.455 & 0.523 \\ \hline
    Sports     & 0.73 & 0.460 & 0.522 \\ \hline
    Politics    & 0.78 & 0.550 & 0.711 \\ \hline
    \end{tabular}
\end{table}

As table \ref{tab:human_agreement} shows, the agreement for all domains is considerably high but not perfect. Agreement is highest---almost 80\%--- for the politics domain. The agreement is lowest in the business domain, $70\%$. The correlations of content-density scores exceed 0.5 and are highly significant ($p<0.001$) for all domains. The high correlations of real-valued scores, especially for the politics and business domains, suggest that the task may be more amenable to annotation and automation as a real-value prediction task rather than as a binary distinction.

Kappa however is relatively low, indicating that the annotation task is rather difficult. To refine our instructions for annotation, we adjudicated all leads for which there was no initial agreement on the label. Both authors sat together, reading the reference lead and each of the leads to be annotated, discussing the reasons why the lead should be labeled content-dense or not. In many cases, the final decision was made by taking into account the domain from which the lead was drawn (i.e. ``there isn't much important information in a sports lead, but it could be considered content-dense in the context of sports news reporting"), as well as the reference lead for the specific genre (i.e. ``the lead is not that content-dense but appears to contain more important facts or reports the news in a more direct style than the reference lead"). We study further the way domain and perception of content density interact in the next section, where independent annotators rated content-density both in in-domain and in domain-independent general settings.\footnote{ As we will shortly see, the classifier is impressively accurate on instances in which the annotators agreed in their initial annotation and quite poorly on the leads that required adjudication. These findings suggest that in future work in may be beneficial to develop a classifier for sentence-level prediction of content-density, which would be helpful for characterizing leads that mix informative and entertaining sentences. Another clear alternative is to develop a classifier to predict that a text is ambiguous in terms of its content-density status.}

Below is an example on whose label the authors initially disagreed. In this lead, the first paragraph is non-informative and the second paragraph is informative, providing partial justification for either overall label.

\vspace{2mm}
\begin{normalsize}
\textit{\textbf{[Example of labelling disagreement]}
Many elderly people are already distressed by the increasing numbers of drugs they are taking, including painkillers and heart medication. Now, those who are also battling depression may be wondering where it all will end.
}

\vspace{-1mm}
\textit{
Last week, researchers at the University of Pittsburgh presented findings from a large government-financed study suggesting that antidepressants are more effective in warding off a recurrence of late-life depression than periodic sessions of interpersonal therapy, a standardized form of talk treatment.
}
\end{normalsize}
\vspace{2mm}

\subsubsection{AMT annotation set}
We also compiled a second set of 600 NYT articles, 150 for each domain. In an attempt to provide more guidance to the annotators, we gave four reference leads for each domain, two as examples of prototypical content-dense leads and two as example of leads that are clearly not content dense. The reference leads for each domain are shown in Appendix A.
The annotators saw the four prototypical leads, as well as a group of leads that they had to annotate. They provided both a categorical label for each target lead (content-dense or not) and a real value score for the degree to which it can be considered content-dense (range between 0 and 100). 

The annotation was partitioned into groups of five leads---an annotator had to label at least five leads and then request more data for annotation, in groups of five. To embed some quality control, one of the five leads in each group is a lead from the dataset annotated by the authors, for which they agreed in independent annotation before the adjudication step. This data allowed us to asses the quality of annotations after problematic annotators were filtered out.

Here we also study the differences in how the content-density of a text would be perceived in-domain and in general setting. For each lead text, two tasks were published separately for labeling content-density in-domain or in general. For the in-domain task, annotators are given domain information (i.e. ``Here are articles drawn from the Sports section of a newspaper...") and the reference leads are selected from that domain. In the general task, workers are not told the domain of the lead and the reference leads were selected without regard to domain.\footnote{The content-dense example was from Business and Science, and the non content-dense from Business and Politics.}

Ten annotators annotated each lead in each of the two conditions.

\paragraph{AMT annotation refinement}
We use two rules to filter out unqualified annotators. We filtered out all annotations by annotators who annotated too quickly or were inconsistent. The first rule is that annotator's average annotation time per task should be longer than 40 seconds. 
 For reference, the average annotation time per task among all annotators is around two minutes.  The second rule is that labeled category and score should be consistent for each lead text. If an annotator labels a lead as content-dense but gives  a very low content-dense score or vice versa, we know something in their understanding of the task is amiss.   
 
There are on average 8 annotators for each item after filtering out unqualified words. For each lead, we use the majority category as the final category label and the average score as the final score label. If there is a tie for a lead, we label it content-dense.

Table \ref{tab:amt_accuracy} shows the agreements and kappas between the majority label from AMT workers and the authors' agreed labels. We compute these only for the in-domain labels because our initial annotation was domain dependent. 

Agreement for the business and sports category is high but only moderate for science and politics. We are unsure about the exact reasons why this is the case. 

\begin{table}[!htb]
    \centering
    \caption{Agreement of embedded baseline leads between AMT workers and authors of the paper.}
    \label{tab:amt_accuracy}
    \begin{tabular}{| c | c | c | }
    \hline
    & Agreement(\%) & Kappa   \\ \hline 
    Business & 92.1 & 0.841  \\ \hline 
    Science   & 86.8 & 0.622  \\ \hline 
    Sports     & 97.3 & 0.947  \\ \hline 
    Politics    & 79.0 & 0.574  \\ \hline 
    \end{tabular}
\end{table}

\begin{table}[!htb]
    \centering
    \caption{Number (and percentage) of content-dense leads annotated by AMT workers for each domain. The same data is annotated with respect to in-domain and general criteria and the statistics for each condition are shown in the first and last column respectively. The two middle columns show the number of leads that changed labels from content-dense(CD) to non content-dense(Non-CD) or vice versa between the in-domain and general condition, broken down according to the direction of the change.}
    \label{tab:amt_labels}
    \begin{tabular}{| c | c | c | c | c |}
    \hline
    & \multirow{ 2}{*}{In-Domain} & \multicolumn{2}{c|}{Label Changes} & \multirow{ 2}{*}{General}   \\ \cline{3-4}
    &&CD $\rightarrow$ Non-CD& Non-CD $\rightarrow$ CD& \\ \hline
    Business &  93 ($62.0\%$) &   8 &  11 &  96 ($66.0\%$) \\ \hline
    Science   &  64 ($42.7\%$) & 16 & 25 &   73 ($48.7\%$) \\ \hline
    Sports     &  76 ($51.1\%$) & 38 &   2 &   40 ($26.7\%$) \\ \hline
    Politics    &  72 ($48.0\%$) &   2 & 53 & 123 ($82.0\%$) \\ \hline 
    \hline
    Overall    & 305 ($50.8\%$) & 64 & 91 & 332 ($55.3\%$) \\ \hline
    \end{tabular}
\end{table}


AMT workers annotated leads in two conditions: in-domain, where the judgements were specific to the domain from which the lead was drawn and general (domain-independent), where a domain was not specified and text from all four domains were randomly mixed in the annotation tasks. Table \ref{tab:amt_labels} shows the number of content-dense leads for each domain for both conditions, along with the number of leads whose labels changed across conditions.  
The first and the forth column correspond respectively to the number (percentage) of content-dense leads among all in-domain and general labels for the same data. The second and third columns show the number of labels that changed their labels from content-dense(CD) to non content-dense(Non-CD) or vice versa, between the domain-dependent and the domain-independent labelling. 

Clearly, the domain context plays a large role in the perception of content density. The change is most clear for the politics and sports domain: in the domain-independent labeling a large number of sports leads, which appeared content-dense for their domain, are considered non content-dense in general. Similarly many of the politics leads considered non content-dense for the standards of the politics domain are considered as such in the domain-independent setting. There are virtually no changes in label in the opposite direction, which conforms to our expectations and provides an additional confirmation of the reasonable quality of the crowdsourced annotations. 

Overall the in-domain annotators have a more balanced number of content-dense and non content dense labels. 

\subsection{Are leads informative?}

In automatic summarization research, the article leads are generally considered to be informative, or content-dense. The beginning of the article is known to be a strong summary baseline \cite{mani02,DBLP:conf/aaai/Nenkova05} and many features for identifying important content in articles are based on overlap with the opening paragraph. Our annotations allow us to directly examine to what extent this general intuition holds across domains of journalistic writing in the New York Times.

Table \ref{tab:amt_labels} shows the number of leads in each domain labeled as content-dense in the manually annotated dataset described above. It is clear that the prevailing assumption that the lead of the articles is always content-dense is not supported in the data we analyze here. 

The majority of articles in the politics domain, which are representative of the data on which large-scale evaluations of summarization system tend to be performed and which focus on specific current events, are indeed content-dense. More than $60\%$ of leads in this domain are labeled as content-dense in the authors' annotation. The trend is similar in the AMT annotations. 

Conforming to intuition, the second largest proportion of content-dense leads is in the business domain. There the articles are often triggered by current events but here is more analysis, humor and creativity. In these leads important information can often be inferred but is not directly stated in factual form. Business leads also tend to have the same labels, regardless of whether they are annotated with respect to the domain standard or in general. For the business domain, only 19 out of 150 labels changed across conditions (cf. first line in Table  \ref{tab:amt_labels}), which corresponds to at least half the rate of label change for any of the other domains.   

In sports the factual information that has to be conveyed is not much and it is embellished and presented in a verbose and entertaining manner. Particularly AMT annotators consider less than a third of the sports leads to be content-dense across domains. In the science journalism section many leads only establish a general topic or an issue, or include a human interest story.  Overall there is only a small portion of science leads labeled as content-dense.

The perception of content density is certainly influenced by the context of the domain. There are 55 politics leads that changed labels from in-domain to the general condition, and 53 of them are changed from non content-dense to content-dense, indicating that in that setting annotators followed their domain bias in deciding the label. Similarly 38 sports in-domain-content-dense leads are non content-dense across domains, but only 2 leads changed in the opposite direction. 

These findings have two important implications for language processing applications and summarization in particular. 

It is unrealistic to expect that all newspaper text has high informational value. Finding valuable content has been addressed as a standalone problem in social media \cite{becker2011beyond} and user generated data \cite{agichtein2008finding} but generally has been ignored in news analysis. 

In addition,  our analysis casts doubt on the practicality of requiring summarization systems to produce summaries of fixed length. Many of the articles with leads that are not content-dense do not discuss even in the body of the article an event readers would consider important. An appropriate summary should simply indicate this, or a summary should not be even attempted. Automatic systems are anyhow not particularly good at summarizing articles that deal with opinion or discussion rather than a specific event \cite{DBLP:conf/acl/NenkovaL08}. In information access applications, tagging the genre of the article as event-centered or not (similar to earlier work in distinguishing opinion pieces from factual reporting \cite{Yu:2003:TAO:1119355.1119372}) may be most helpful, with preview snippet summaries produced only for the event-centered articles. 

\subsection{Classifier evaluation}
Here we evaluate the combined two-layer classifier trained on heuristically labeled data on the manual annotations. 
Note that the manual annotated leads are used for evaluation only, no additional training is performed at this stage.

\paragraph{Baselines} Following the assumption prevailing in summarization research that the lead of the article is always content-dense, the first baseline (Baseline-1) always  considers the lead of the article content-dense. 

The second baseline (Baseline-2) is established based on the length of the entire news article, not only the lead. The intuition is that longer articles may have uninformative leads designed to draw the reader into the subject while short articles need to start out with a more focused presentation of the event so are likely to have an content-dense lead. 
We train a L2-regularized logistic regression model based on this single feature. As table \ref{tab:res_human} shows, the single feature classifier achieve reasonable accuracy of  $68\%$ for the science domain.

\paragraph{Classification results on the basic set}
Table \ref{tab:res_human} shows the results from applying the domain-dependent and the general domain-independent models on the basic human annotation set. Accuracy  computed against each of the two individual annotators is shown in the last two columns. Sports and politics domains have higher prediction accuracies on the data labeled by the first annotator, and business and science domains have higher prediction accuracies for the second annotator's labels. Also the prediction accuracies have smaller variances on the data labeled by the first annotator, between $78\%$ for the politics domain and $74\%$ for science the domain, compared with the accuracies on data labeled by the second annotator, between $87\%$ for business and $71\%$ for sports. Overall however the prediction accuracy on the final combined data, after disagreements have been adjudicated,  is highest, demonstrating that the adjudication procedure did lead to more internally consistent labels. 
As in the heuristically labeled data, recognition accuracies are higher for the business and science domains ($83\%$) and lower for the sports and politics domains (around $80\%$). 

We also evaluate the prediction accuracy separately on the subsets of the data for which the two annotators agreed on the label in the first stage of independent annotation, corresponding to the presumably clear-cut cases, and those for which adjudication was needed. Clearly, the classifier captures characteristics of content-dense leads quite well. The accuracy on the subset of the data for which the annotators agree is much higher than that for individual annotators, indicating that when the text has mixed characteristics leading to disagreement in annotation, it is more likely that the classifier makes more errors as well. 

On the agreed subset---marked with the same label by both annotators during independent annotation---accuracies are around $90\%$ for the business and science domains, $80\%$ for sports and politics domains. 

The classifier accuracies are much higher than the baselines for all domains. 

\begin{table}[!htb]
    \centering
    \caption{Binary classification accuracies(\%) on basic human annotated datasets for models trained on heuristically labeled data. }

    \label{tab:res_human}
    \begin{tabular}{ c | c c c | c c }
        \hline
        \textbf{Business} & Combined & Agreed & Adjudicated & Anno\_1 & Anno\_2  \\ \hline
        Domain model   & \textbf{83} & \textbf{94.3} & \textbf{56.7} & \textbf{75} & \textbf{87} \\ 
        Overall model    & 79             & 91.4             & 50.0             & \textbf{75}  & 83 \\ 
        Baseline-1         & 53             & 52.8             & 53.3             & 47              & 57 \\ 
        Baseline-2         & 60             & 65.7             & 46.7             & 58              & 64  \\ \hline
        \hline
        \textbf{Science} & Combined & Agreed & Adjudicated & Anno\_1 & Anno\_2  \\ \hline
        Domain model   & \textbf{83} & \textbf{89.2} & \textbf{65.4} & \textbf{77} & 80 \\ 
        Overall model    & 81             & \textbf{89.2} & 57.7             & 71              & \textbf{86} \\ 
        Baseline-1         & 37             & 31.1             & 53.8             & 45              & 27 \\
        Baseline-2         & 68             & 69             & \textbf{65.4} & 62              & 65 \\ \hline
        \hline
        \textbf{Sports} & Combined & Agreed & Adjudicated & Anno\_1 & Anno\_2 \\ \hline
        Domain model   & \textbf{78} & \textbf{80.8} & 70.3             & \textbf{74} & \textbf{71} \\ 
        Overall model    & 75             & 75.3             & \textbf{74.1} & 69              & 68 \\ 
        Baseline-1         & 49             & 46.5             & 55.6             & 45              & 50 \\ 
        Baseline-2         & 63             & 70             & 51.9             & 63              & 66 \\ \hline
        \hline
        \textbf{Politics} & Combined & Agreed & Adjudicated & Anno\_1 & Anno\_2 \\ \hline
        Domain model   & 78             & \textbf{83.3}  & 59.1            & \textbf{78} & 74 \\ 
        Overall model    & \textbf{80} & \textbf{83.3} & \textbf{68.2} & 76             & \textbf{76} \\ 
        Baseline-1         & 61             & 60.3              & 63.6             & 55             & 61 \\
        Baseline-2         & 51             & 55              & 36.4             & 55             & 53 \\ \hline
        \hline
    \end{tabular}
\end{table}

\begin{table}[!htb]
    \caption{Correlation between predicted probabilities and human annotated scores. All correlations are highly significant with $p<0.001$. }
    \label{tab:corr_human}

    \begin{tabular}{| c | c  c | c  c |}
    \hline
    & \multicolumn{2}{c|}{\textbf{Annotator\_1} } & \multicolumn{2}{c|}{ \textbf{Annotator\_2} } \\ \cline{2-5}
    & Domain Models & Overall Models & Domain Models &Overall Models  \\ \hline
    Business & 0.621              & \textbf{0.647} & 0.797             & \textbf{0.810} \\  
    Science   & \textbf{0.575} & 0.546              & 0.711             & \textbf{0.758} \\  
    Sports     & \textbf{0.590} & 0.575              & \textbf{0.588} & 0.582 \\  
    Politics    & \textbf{0.658} & 0.629              & \textbf{0.609} & 0.592 \\  
    \hline
    \end{tabular}
     \centering
\end{table}

Table \ref{tab:corr_human} shows the correlations between the classification score from the final classifier and the real-value score of content-dense by the two annotators. All correlations are highly statistically significant. In line with what we have seen in the analysis of other results, the correlation is the highest for the business domain.

\begin{figure}[!h]
    \centering
    \begin{subfigure}[b]{0.49\textwidth}
            \centering
            \includegraphics[width=\textwidth]{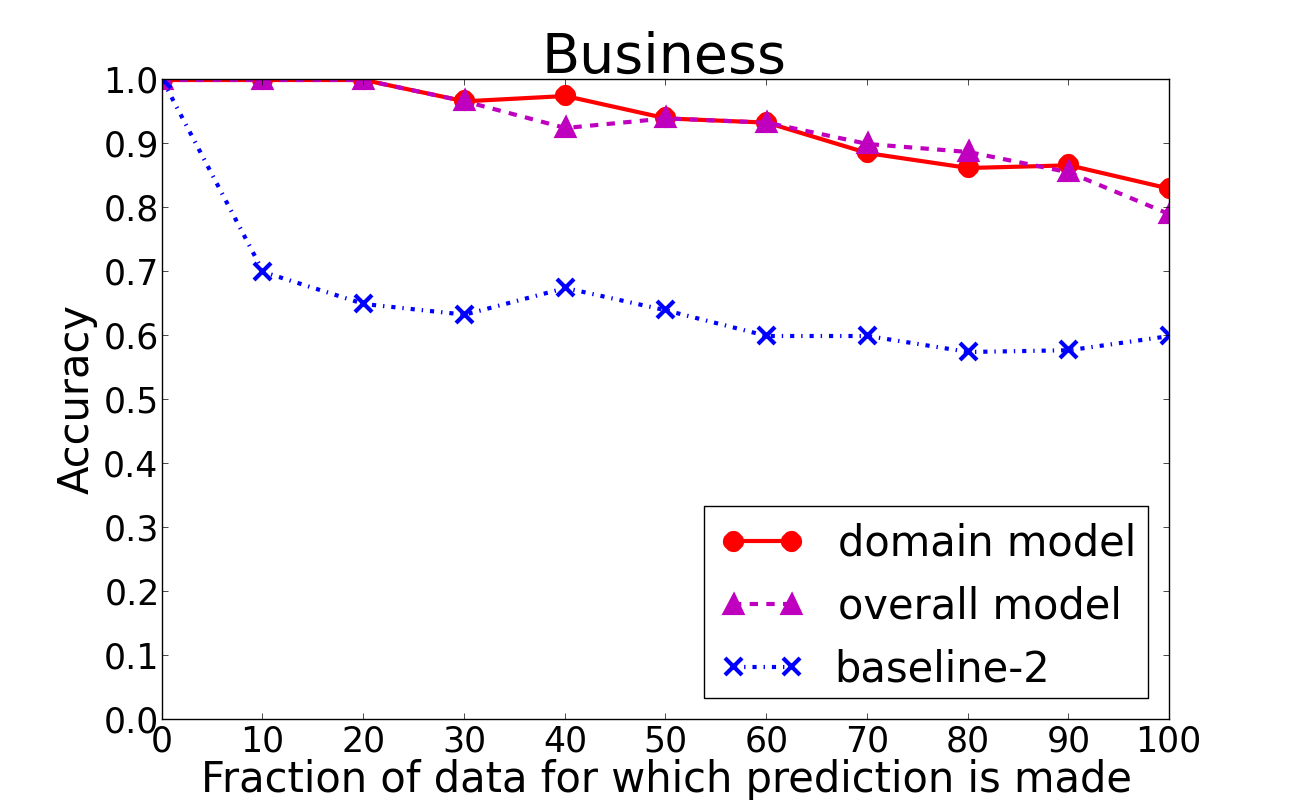}
    \end{subfigure}
    \begin{subfigure}[b]{0.49\textwidth}
            \centering
            \includegraphics[width=\textwidth]{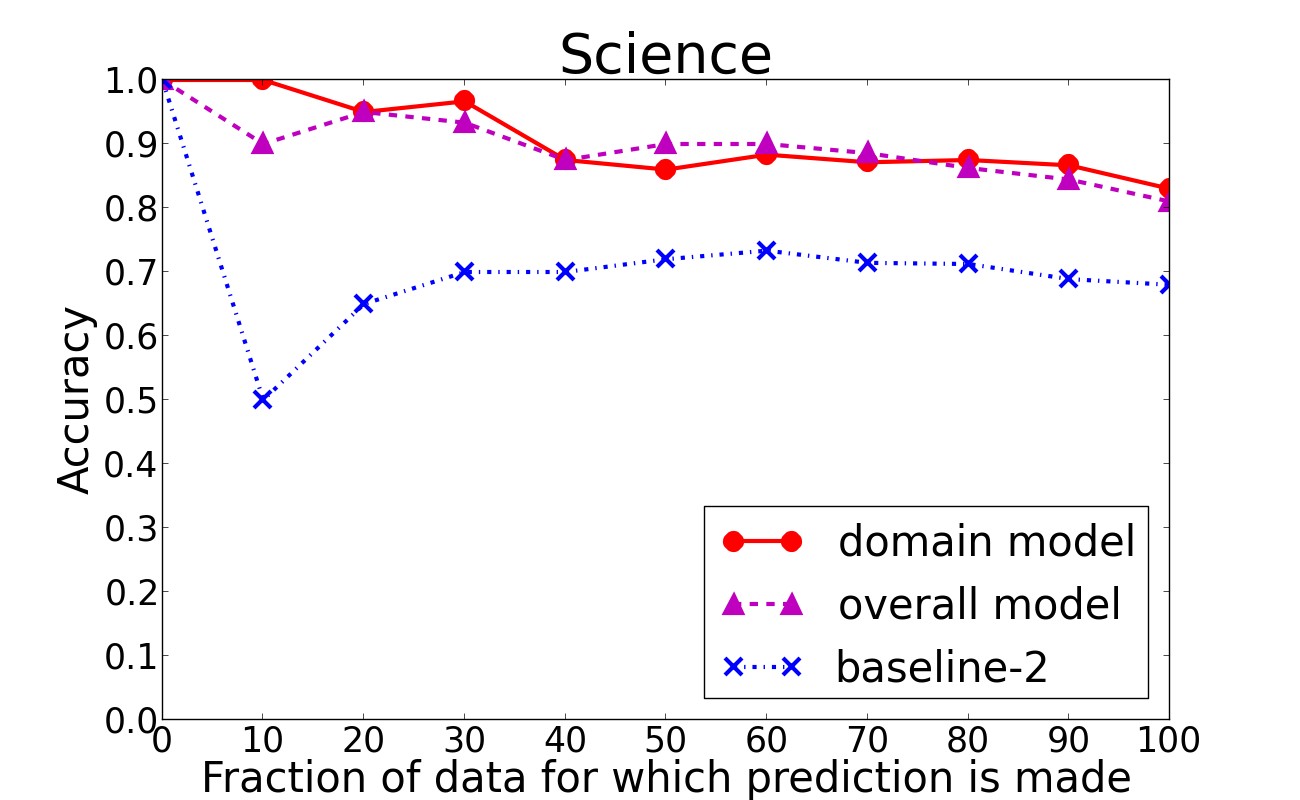}
    \end{subfigure}
    \begin{subfigure}[b]{0.49\textwidth}
            \centering
            \includegraphics[width=\textwidth]{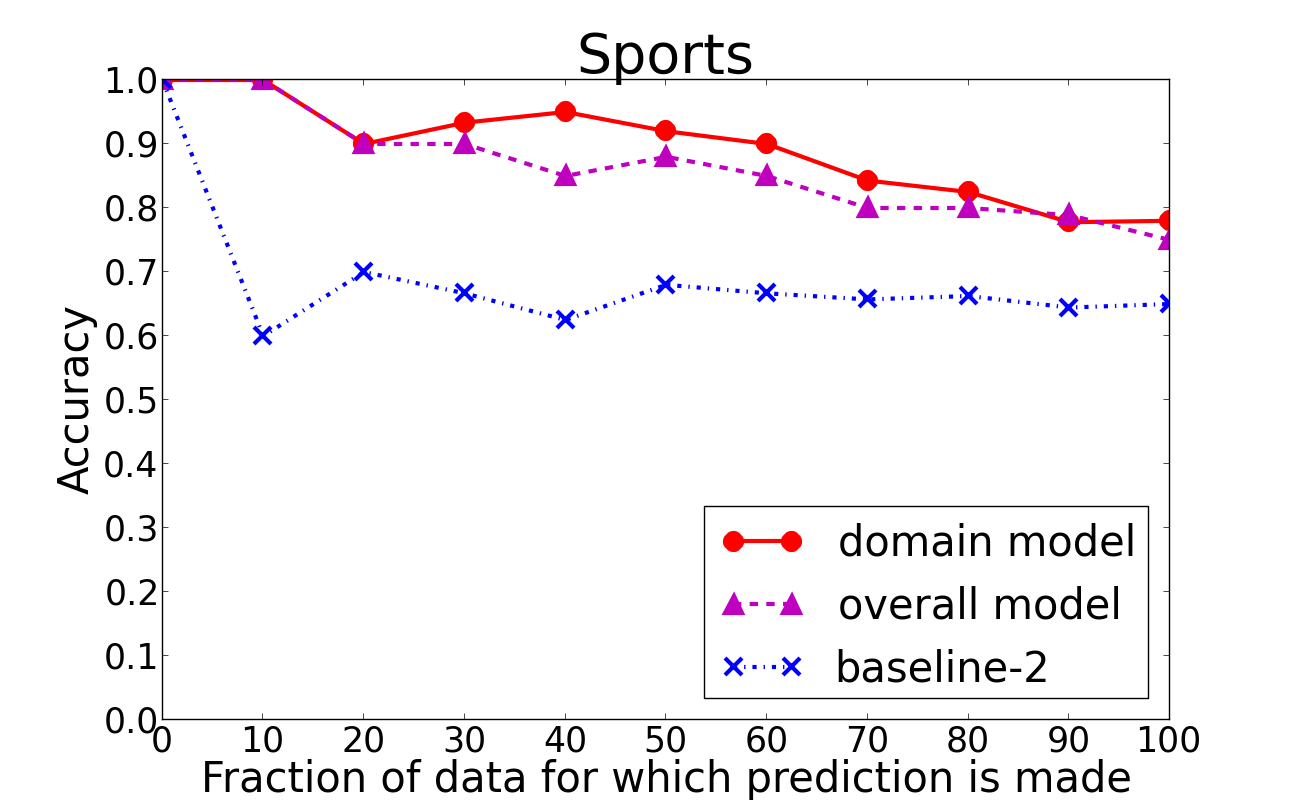}
    \end{subfigure}
    \begin{subfigure}[b]{0.49\textwidth}
            \centering
            \includegraphics[width=\textwidth]{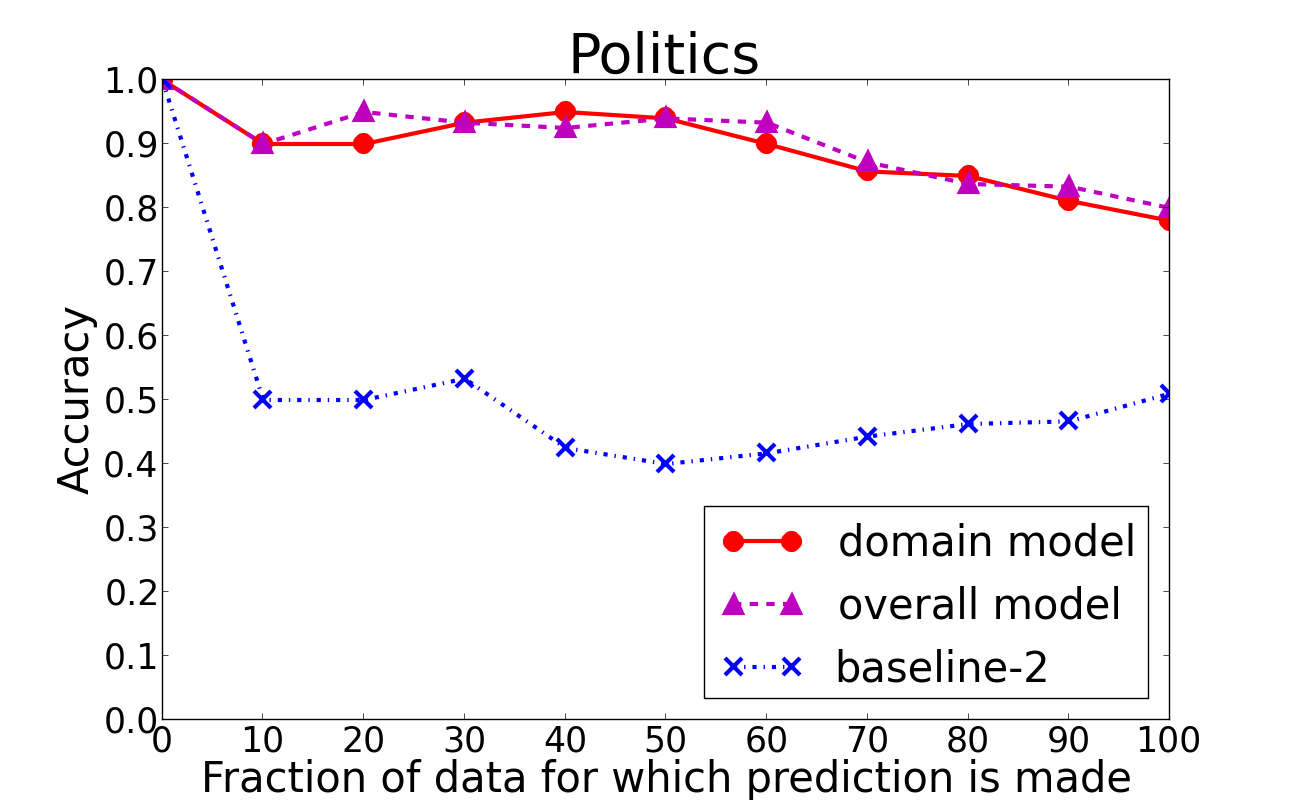}
    \end{subfigure}
    \caption{Prediction accuracy based on probability ranking on basic human annotation set. \textbf{[Top Left]} Business,  \textbf{[Top Right]} Science,  \textbf{[Bottom Left]} Sports,  \textbf{[Bottom Right]} Politics}
    \label{fig:pr_human}
\end{figure}

Similarly we compute the prediction accuracy stratified according to the classifier confidence in that prediction. Figure \ref{fig:pr_human} shows the plot on all four genres. The accuracy of high confidence predictions is much higher than the overall accuracy. The "article length" baseline, however, has lower accuracy in its high confidence predictions.    

\paragraph{Classification results on the AMT annotations}
Table \ref{tab:res_amt} shows the accuracy  of the domain-dependent and the general domain-independent models on the AMT annotations. As in previous tables, row 1 and 2 represent the results from domain models and the domain-independent models respectively. Rows 3 to 4 show results for the two baselines. Our classifiers outperform the baselines by a large margin except for politics in the domain-independent labels, where the baseline that considers all leads to be content dense works best.
Overall however, the results show that the baseline of assuming all leads are content-dense performs poorly and the proposed approaches significantly improve the accuracies. 

Comparing the accuracies of prediction for data drawn from the same newspaper section, it is evident that business and science have the most stable prediction and the accuracy of the domain-dependent and the domain-independent classifiers does not differ much on these subsets of the test data. 
The classifier trained on domain-independent labels achieves  $78.0\%$ accuracy  on the domain-specific labels, in which the annotators were explicitly told the news section from which the article was drawn and used this information in judging if the lead is content-dense or not. This accuracy is less than $2\%$ lower than the prediction on domain-independent labels. Similarly in the science domain the difference in performance for the two types of labels is as  low as $2.0\%$. In stark contrast, there is large difference in performance for the in-domain and domain-independent labels for sports and politics, where the difference between the two reaches 10\%. 

The crowdsources annotations were performed both in-domain (judging the content density with respect to the expectation for the given domain, politics for example) and in domain independent setting. Here domain models  are on average worse than the domain independent models. For the sports domain, training a domain-specific classifier helps most in improving the detection of content-dense sports leads but for the other domains the advantage is less clear. This result is reassuring. If the domain models were clearly superior, one would have needed accurate domain predictors for practical applications. The analysis presented here demonstrates that a domain-independent classifier may be sufficient for many applications.


The accuracies of the domain models drop considerably compared to their respective accuracies on the author-annotated set. For example, there is around $8.0\%$ drop in the politics domain. There are several possible reasons for this difference. The articles in the initial set that the authors annotated were selected only from the articles published in 2005 and 2006 while  the AMT set is selected from the entire NYT dataset from 1987 to 2007. The annotation instructions also differed for the two sets. The  AMT annotators were presented with  prototypical content-dense and non content-dense leads as references, while the authors had only one lead in the middle of the range  of content-density as reference. Finally, the general domain-independent classifier on average works best, predicting both the in-domain and general labels in the test set better than the domain-dependent classifiers. This trend indicates that AMT workers were likely more influenced by general domain expectations when labeling the data. It is plausible that domain-dependent annotation is requires more detailed instructions that are not as readily passed on in the crowdsourced setting. 

\begin{table}[!htb]
    \centering
    \caption{Binary classification accuracies($\%$) on AMT annotated datasets for models trained on heuristically labeled data. }

    \label{tab:res_amt}
    \begin{tabular}{ c | c c c c c }
        \hline
        \textbf{In-domain} & Business & Science & Sports & Politics & Average \\ \hline
        Domain model   & 76.0             & 72.7             & \textbf{73.3} & \textbf{70.7} & 73.2 \\ 
        General model    & \textbf{78.0} & \textbf{76.7} & 70.0            & 70.0              & \textbf{73.7} \\ 
        Baseline-1         & 62.0             & 42.7             & 51.1             & 48.0              & 55.0 \\ 
        Baseline-2         & 58.0             & 64.7             & 62.7             & 52.7              & 59.5 \\ \hline
        \hline
        \textbf{Domain indep.} & Business & Science & Sports & Politics & Average  \\ \hline
        Domain model   & \textbf{79.3} & 73.3             & 80.0             & 76.7              & 77.3 \\ 
        General model    & 77.3             & \textbf{78.7} & \textbf{82.0} & 77.3              & \textbf{78.8} \\ 
        Baseline-1         & 66.0             & 48.7             & 26.7              & \textbf{82.0} & 55.9 \\ 
        Baseline-2         & 62.7             & 66.7             & 68.0              & 50.7             & 62.0 \\ \hline
        \hline
    \end{tabular}
\end{table}

We further compute the correlations coefficients between predicted probabilities and average scores annotated by AMT workers. The results are shown in table \ref{tab:corr_amt}. Domain models have better correlations than general models in three of domains for domain dependent (in-domain) labels, but with small absolute difference in correlation.  The domain-independent models are much better in predicting content-dense in the general, domain-independent condition. 
All correlations are highly significant, range from 0.577 to 0.661 against in-domain labels and from 0.602 to 0.730 against domain-independent labels. As in the binary prediction task, the domain-independent label appear to be easier for the system to predict.
The correlation coefficients are in line with our intuition and much closer to the numbers we have seen based on the basic author-annotated set (shown in table \ref{tab:corr_human}). This trend implies that predicting content-density in terms of real-value scores may be more suitable for this task. 

\begin{table}[!htb]
    \caption{Correlation between predicted probabilities and average scores annotated by AMT workers in the domain specific and general condition. All correlations are highly significant with $p<0.001$. }
    \label{tab:corr_amt}
    \begin{tabular}{| c | c c | c c |}
    \hline
     & \multicolumn{2}{c|}{\textbf{In-Domain Labels} } & \multicolumn{2}{c|}{ \textbf{Domain Independent Labels} } \\ \cline{2-5}
    & Domain Models & General Model & Domain Models & General Model \\ \hline
    Business & 0.602              & \textbf{0.614} & 0.713             & \textbf{0.730} \\  
    Science   & \textbf{0.661} & 0.646              & 0.652             & \textbf{0.690} \\  
    Sports     & \textbf{0.600} & 0.577              & \textbf{0.619} & 0.602 \\  
    Politics    & \textbf{0.616} & 0.615              & 0.652             & \textbf{0.668} \\  
    \hline
    \end{tabular}
     \centering
\end{table}

\begin{figure}[!h]
    \centering
    \begin{subfigure}[b]{0.49\textwidth}
            \centering
            \includegraphics[width=\textwidth]{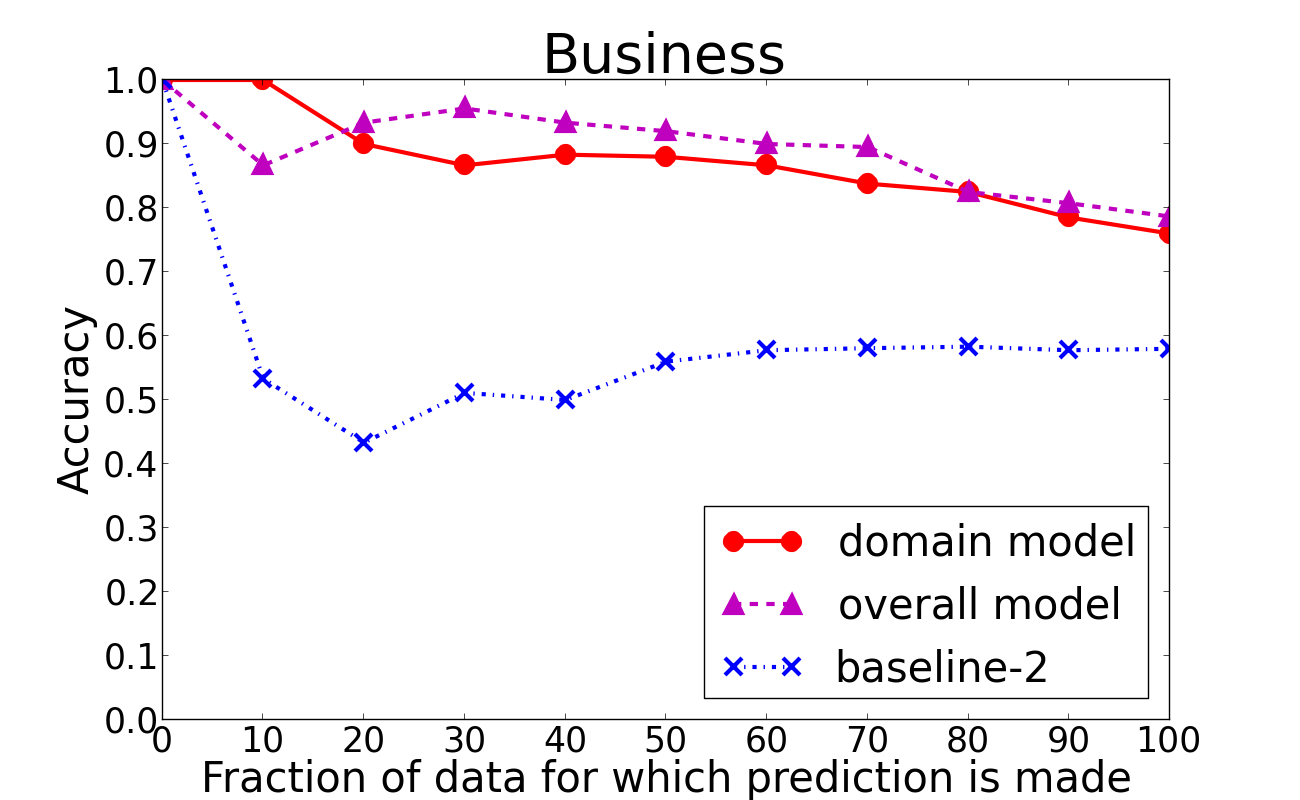}
    \end{subfigure}
    \begin{subfigure}[b]{0.49\textwidth}
            \centering
            \includegraphics[width=\textwidth]{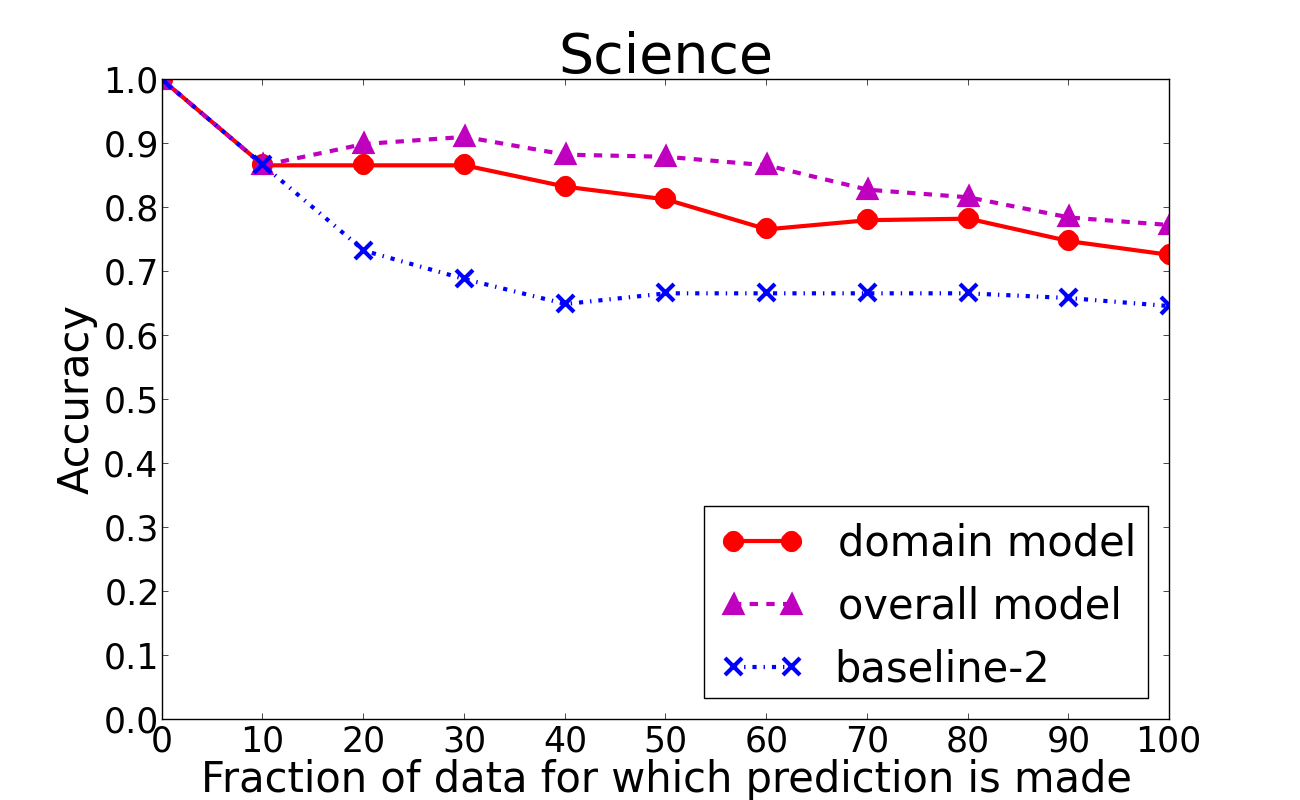}
    \end{subfigure}
    \begin{subfigure}[b]{0.49\textwidth}
            \centering
            \includegraphics[width=\textwidth]{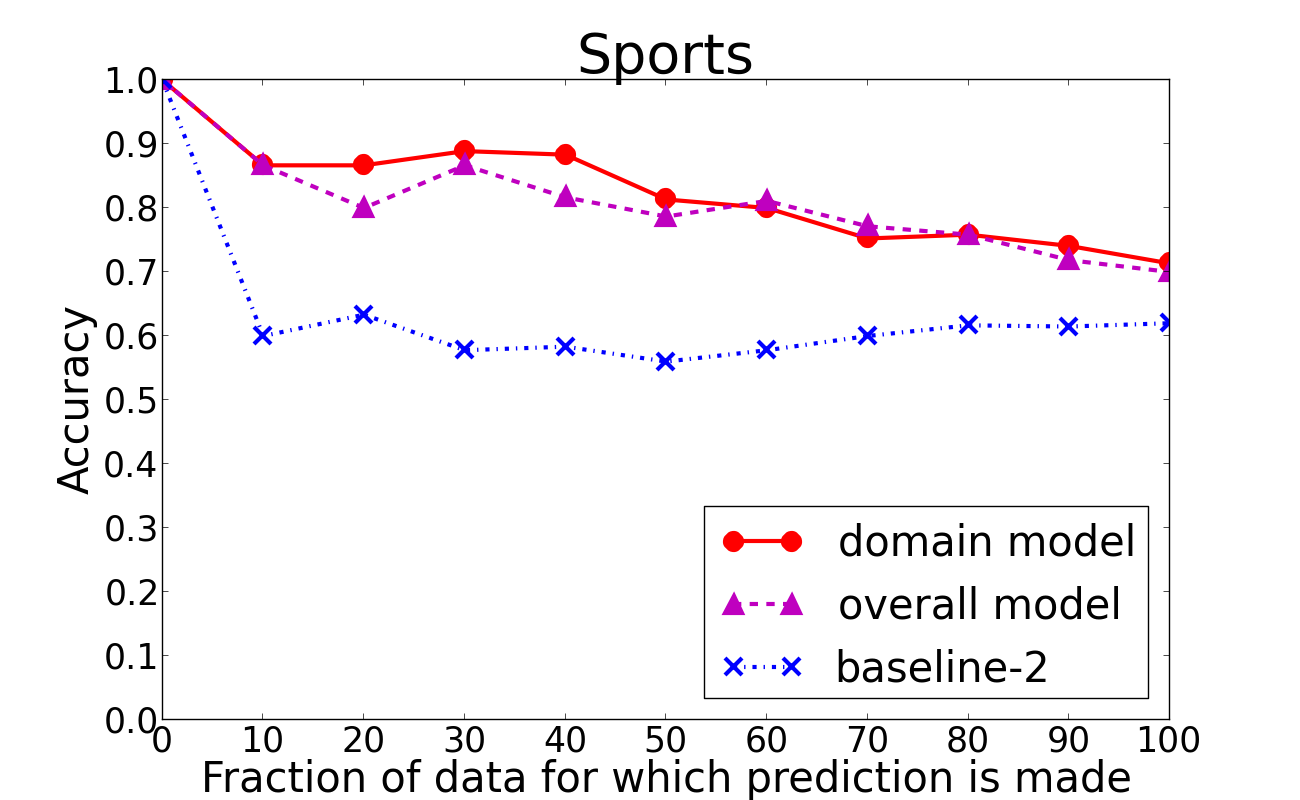}
    \end{subfigure}
    \begin{subfigure}[b]{0.49\textwidth}
            \centering
            \includegraphics[width=\textwidth]{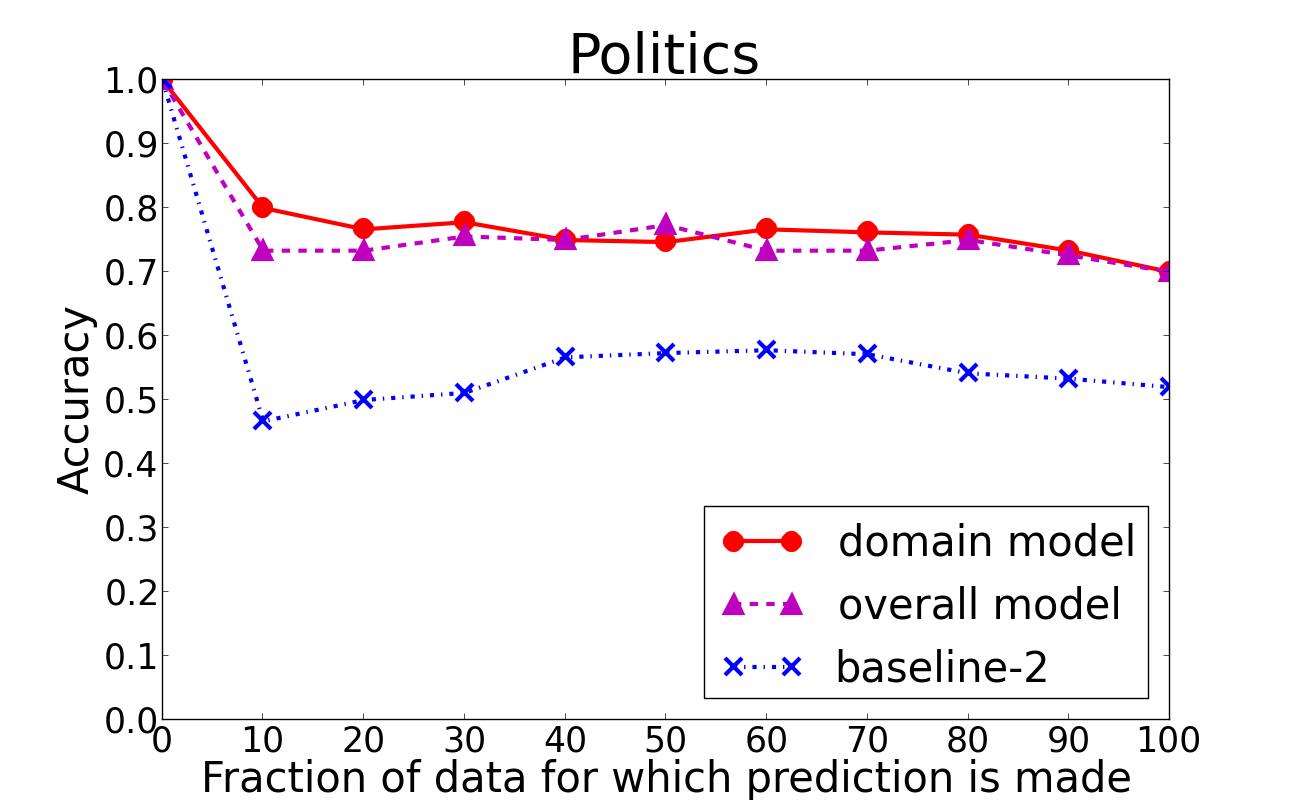}
    \end{subfigure}
    \caption{Prediction accuracy based on probability ranking on AMT annotated data. The x axis represents the percentile of data used to calculate the accuracy according to the predication confidence. (In-domain): \textbf{[Top Left]} Business,  \textbf{[Top Right]} Science,  \textbf{[Bottom Left]} Sports,  \textbf{[Bottom Right]} Politics}
    \label{fig:pr_amt}
\end{figure}

\begin{figure}[!h]
    \centering
    \begin{subfigure}[b]{0.49\textwidth}
            \centering
            \includegraphics[width=\textwidth]{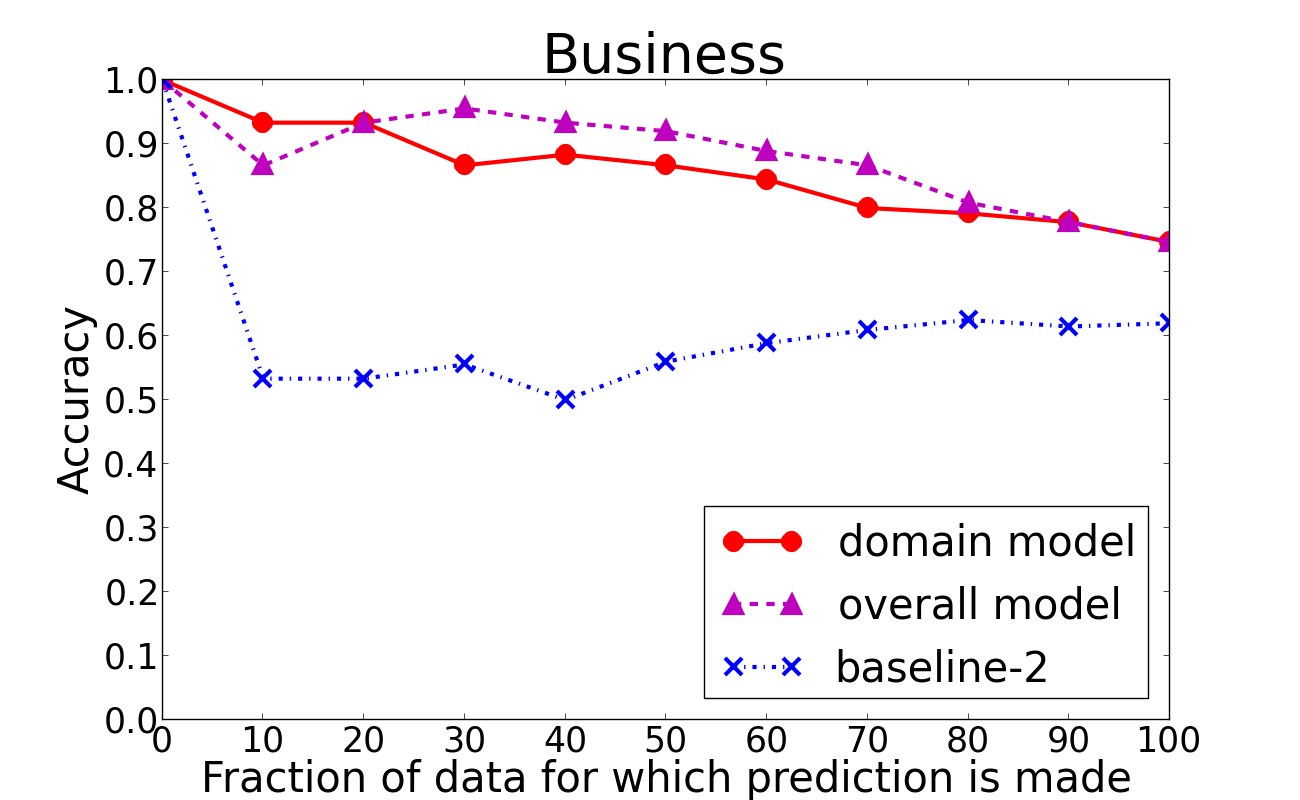}
    \end{subfigure}
    \begin{subfigure}[b]{0.49\textwidth}
            \centering
            \includegraphics[width=\textwidth]{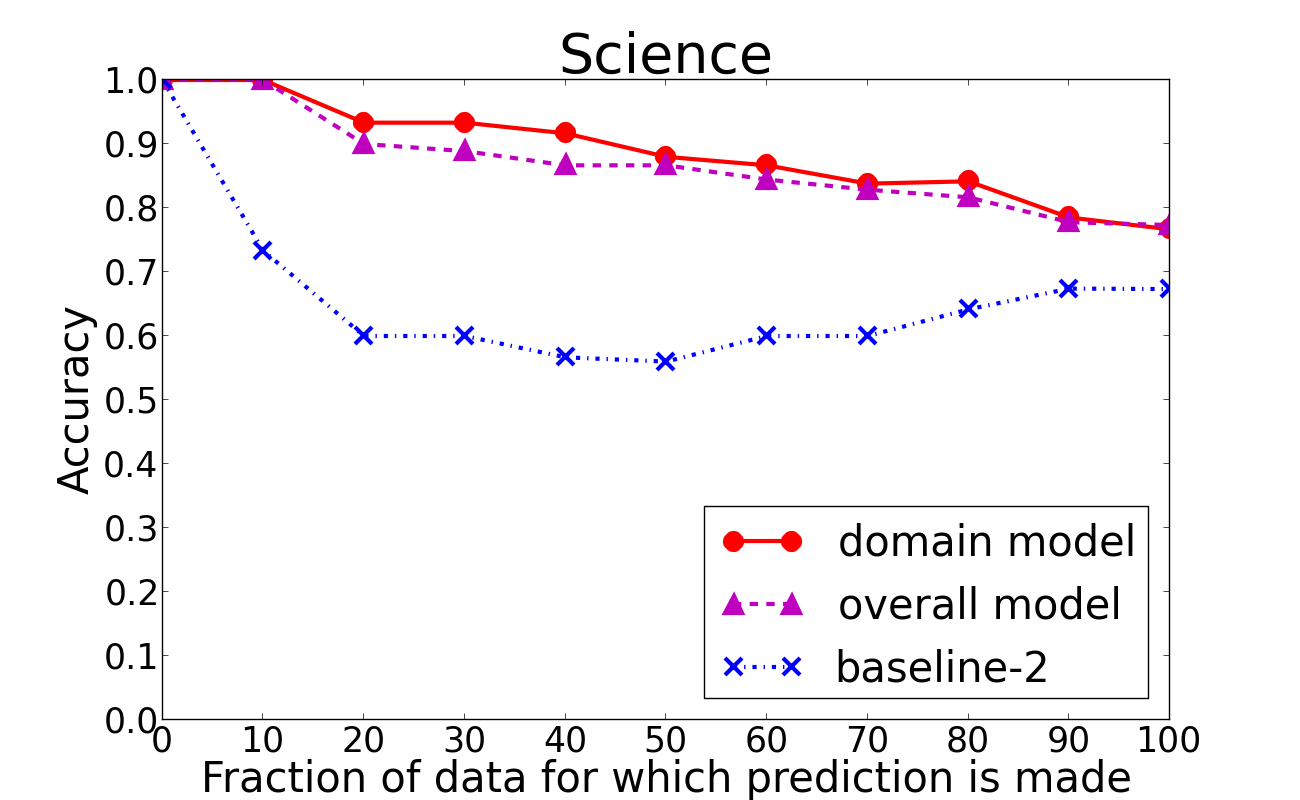}
    \end{subfigure}
    \begin{subfigure}[b]{0.49\textwidth}
            \centering
            \includegraphics[width=\textwidth]{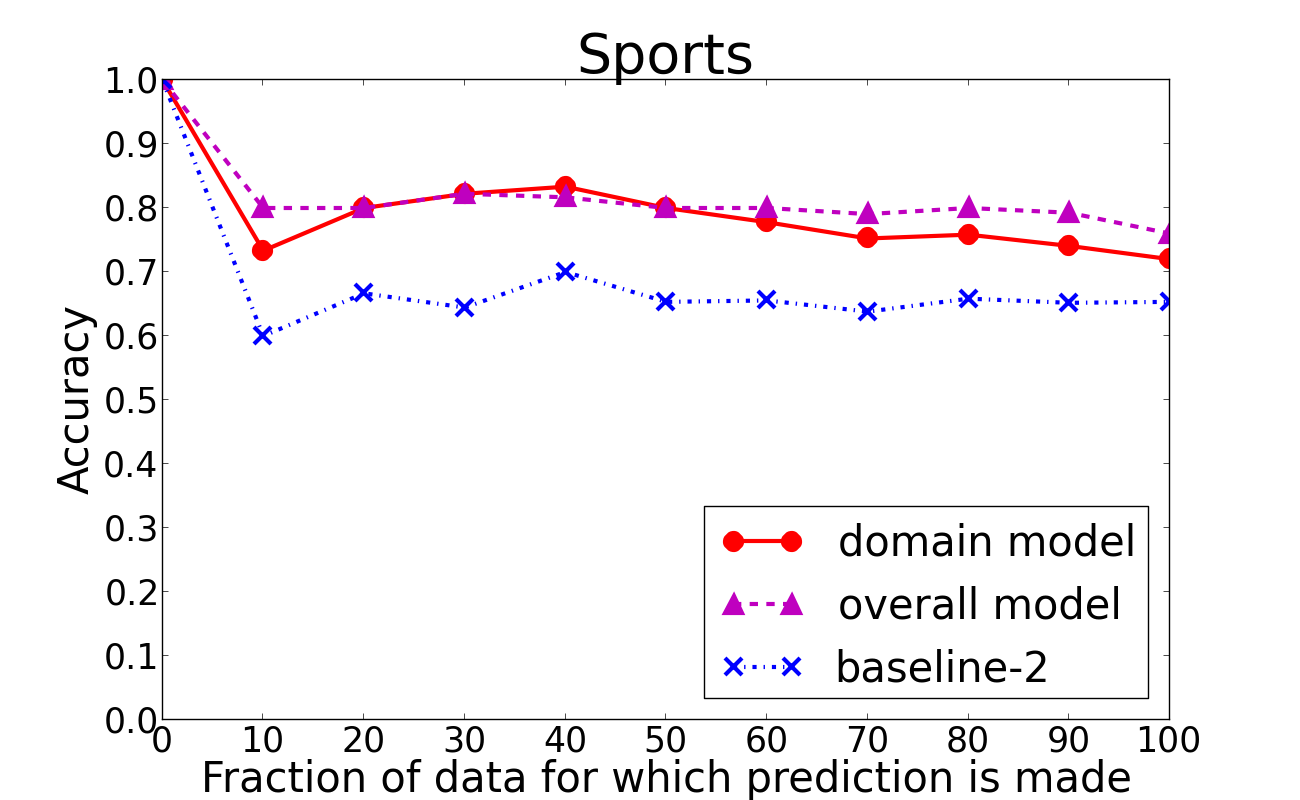}
    \end{subfigure}
    \begin{subfigure}[b]{0.49\textwidth}
            \centering
            \includegraphics[width=\textwidth]{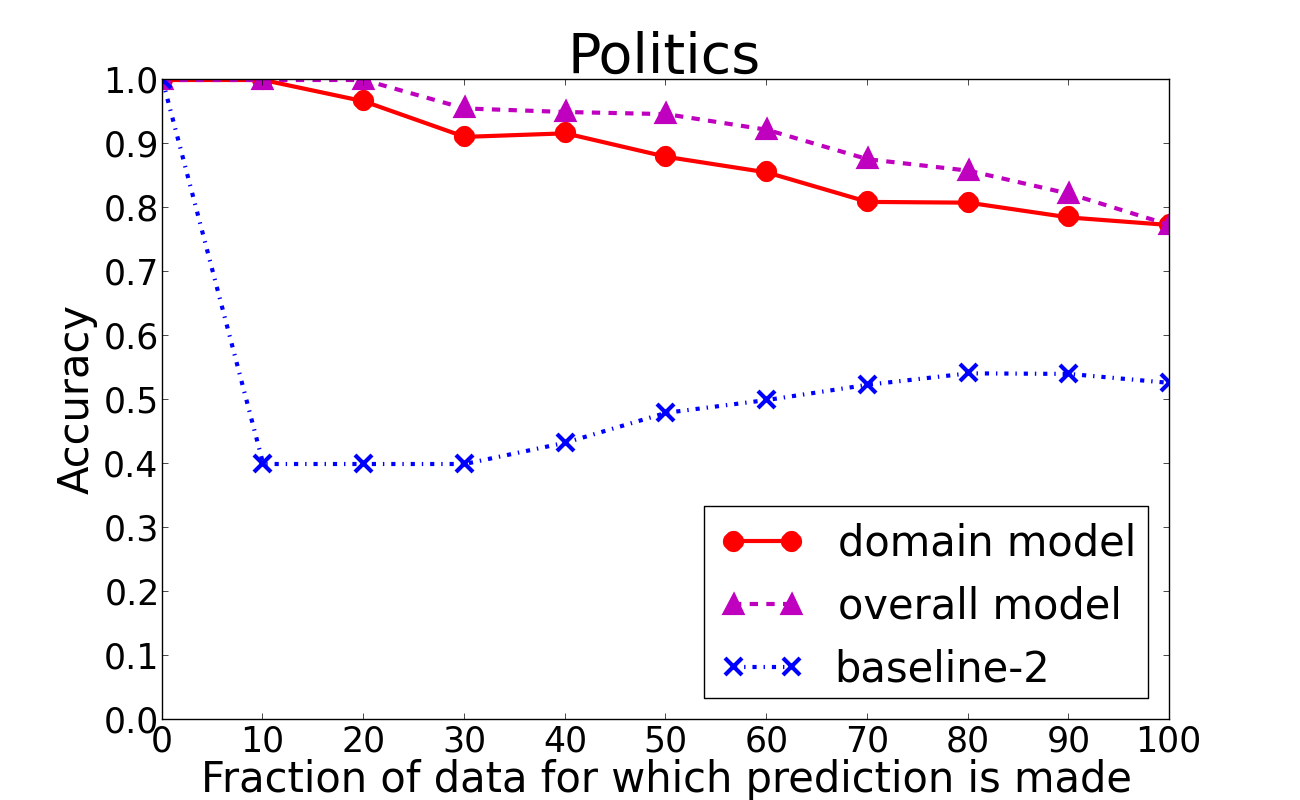}
    \end{subfigure}
    \caption{Prediction accuracy based on probability ranking on AMT annotated data. The x axis represents the percentile of data used to calculate the accuracy according to the predication confidence. (Domain independent): \textbf{[Top Left]} Business,  \textbf{[Top Right]} Science,  \textbf{[Bottom Left]} Sports,  \textbf{[Bottom Right]} Politics}
    \label{fig:pr_amt_cr}
\end{figure}

For the AMT annotated test set, we also compute the prediction accuracy stratified according to percentiles of data ranked by the classifier confidence in that prediction. Figures \ref{fig:pr_amt} and \ref{fig:pr_amt_cr} show the plots on all four domains for the two types of annotated labels (domain-specific or domain-independent). Again, the accuracy of high confidence predictions is much higher than the overall accuracy. The article length baseline, however, has much lower accuracies.   

\section{Recognizing better summaries}
So far we have demonstrated that detector of content-denisty can be developed using heuristically labeled data and that it can achieve respectable accuracy in intrinsic evaluation on human-labeled leads. Ultimately however the goal would be to integare the content-denisty prediction in information seeking applications such as summarization and news browsing. 
Testing the impact of the content-density prediction in such extrinsic evaluations will be the main focus of future work.

Here, however, we show a feasibility study to verify the potential for development of more informed summarization methods that exploit the concept of content density.  Specifically, we demonstrate that the content-density detector is able to recognize when an automatic summary of a single news article is better than the lead of the article. This is an important open problem in summarization, where the lead paragraph baseline is very strong and few systems outperform it \cite{Over2007}. Moreover, all of the proof-of-concept experiments from the previous section were performed on data drawn from the NYT. We would like to verify that the usefulness of the prediction remains when data from other sources is considered. 

Motivated by these goals, we perform our experiments on detecting if a machine summary is more informative than the lead on two datasets: NYT and on data from the Document Understanding Conference, which has data from a variety of sources. 

We randomly selected 400 articles with manual summaries from the NYT, 100 for each genre.
We generated automatic summaries for the article using two systems.
The first system, LeadSumm, is the strong lead baseline which picks the first 100 words as the summary. 
The second system is IcsiSumm \cite{gillick2009scalable}, which is one of the state-of-the-art multi-document summarization systems \cite{gillick2009global,berg2011jointly}.

We performed human evaluation to determine which of the two summaries in the pair (LeadSumm and IcsiSumm) is better.
We asked annotators to first read the manual summary from NYT, then read the two summaries generated by the systems. 
Then we asked the annotators to indicate which of the two system summaries covers better the information expressed in the NYT goldstandrd. Thel were also provided with an option to indicate that the two system summaries cover the information expressed in the goldstandard equally well. 

The flow between the sentences in the LeadSumm summaries is better than those in the automatic summaries because this is a snippet of professionally written discourse. Here our goal is to study the content-density of the two summaries, independently of the linguistic quality of the summary which we know favors the lead system. For this reason, we randomized the order of the sentences in both of the LeadSumm and IcsiSumm summaries. The order of presenting the LeadSumm and IcsiSumm to annotators is also randomized during judgement collection.

The tasks are published on Amazon Mechanical Turk (AMT) and each task is assigned to 10 annotators, with one summary per task/HIT.
IcsiSumm generated empty summaries for 77 out of the 400 randomly selected articles and two of its summaries were identical to the lead baseline. 
We removed those tasks so there are total 323 tasks are published. 
The majority vote of the 10 annotators is used as the final label.
The human annotations are used as ground truth in the following steps. 

Next we apply the content-dense detector on the generated IcsiSumm and LeadSumm to get content-dense probability scores for each.
The summary with higher content-dense score is predicted to be the better summary. As expected from prior manual shared task evaluations, LeadSumm is better then IcsiSumm for most of the articles.


The confidence in the prediction that one summary is better than another is controlled by the content-dense score difference. The larger the difference between the content-density scores of the IcsiSumm and LeadSumm is, the more confident we can be that the summary is indeed better. We track how the summarization performance varies with the difference in content-density scores. In cases when the difference between the two content-denisty scores is lower than a set value, we consider that the lead summary is better.   
By defining the $score\_difference = score_{IcsiSumm} - score_{LeadSumm}$,
we cutoff the evaluations samples by $score\_difference$ compute the metrics for different cutoff levels. 


Table \ref{tab:NYT_comp_acc} shows the results of detecting when IcsiSumm produces better summaries than the lead baseline on the NYT articles. This is equivalent to a combination system which uses lead summaries unless it is confident that the automatic summary is better, in which case it uses the IcsiSumm summary. 
Each show shows the metrics values for a cutoff.
The first column represents the cutoff threshold. 
The second and third columns show the number of total samples and number of positive samples within this cutoff repectively.

The last row of the table shows the total number of judgements and the number of times assessors indicated that a given summary---either the lead or that produced by the ICSI summarizer---provides more information about the described event. The lead was better for 59\% of the test articles; the ICSI summarizer produced the more informative summary for 34\% of the test articles. The two summaries were considered equally informative in the rest of the cases. 

Clearly, as expected from past DUC evaluations, the lead baseline summary is better than the automatic summarizer. However even an extractive summarizer can significantly outperform the lead baseline if we had a reliable way in which to predict when an alternative summary would be more informative; this could improve one out of each three summaries produced by the summarizer.

In Table \ref{tab:NYT_comp_acc} we also show the performance of a combination system using content-density scores to decide which of the two available summaries is superior. Whenever at least some threshold is used to decide when the automatic summary is better, the combination system's output is preferred by the assessors considerably more often than the output for the lead baseline. Particularly for thresholds between 0.1 and 0.4, the output of the combination system is preferred between 64 and 66\% of the time, compared to the 58.5\% for the lead baseline. These improvements would be statistically significant according to a binomial test with expected probability of producing better summary of 0.585, corresponding to the human preference for the baseline lead summaries.

\begin{table}[!htb]
    \centering
    \caption{Performance of combination system with different cutoffs on NYT articles. The last column shows the number (percentage) of correct predicted samples of the combination system. }
    \label{tab:NYT_comp_acc}
    \begin{tabular}{c | c | c | c | c | c || c }
    \hline
    & \multirow{2}{*}{cutoff} & \multirow{2}{*}{\# of samples}
        & \multicolumn{4}{c}{Human Judgement} \\
    \cline{4-7}
    &   &  & Icsisumm & Tie & Leadsumm & Combination System \\
    \hline
    & 0.5 & 18  & 14  & 0 & 4  & 199 (61.6$\%$) \\    
    & 0.4 & 29  & 23  & 1 & 5  & 207 (64.1$\%$) \\    
    & 0.3 & 42  & 32  & 2 & 8  & 213 (65.9$\%$) \\    
NYT & 0.2 & 54  & 39  & 2 & 13 & 215 (66.6$\%$) \\    
    & 0.1 & 79  & 47  & 4 & 28 & 208 (64.4$\%$) \\    
    & 0   & 179 & 78  & 7 & 94 & 173 (53.6$\%$) \\    
    \cline{2-7}
    & All & 323 & 109 (33.7\%) & 25 & 189 (58.5\%) & N/A \\   
    \hline
    \end{tabular}
\end{table}
    


Next we verify that the proposed model works with reasonable accuracy on sources other than the NYT.
We run the same LeadSumm and IcsiSumm systems on the DUC dataset~\cite{Over2007}.
We only perform the experiments on the data from DUC2002, which is the last year NIST provides single-document human summaries. 

There are total of 533 articles from various sources in DUC2002, including Associated Press (AP), Wall Street Journal (WSJ), Los Angeles Magazine (LA), FT Magazine (FT), San Jose Mercury News (SJMN) and Foreign Broadcast Information Service Daily Reports (FBIS). AP is a newswirse service, providing high-quality news reporting  used by many media outlets. By the nature of newswire services, the AP articles (and leads) are expected to be contain a larger proportion of content-dense texts. The other article sources are drawn from newspapers, so are much more likely to include leads that are not content-dense.

By filtering out those articles have no IcsiSumm generated or the two generated summaries are identical.
Again, we exclude from considerations articles for which IcsiSumm did not generate a summary or for which IcsiSumm produced a summary consisting of the lead of the article. After filtering these, we obtain  493 articles for the evaluation. 

As with the NYT experiment, we use AMT to obtain judgements about which of the two summaries of the article is better.
All the annotation settings are exactly the same with the NYT annotations described above.

\begin{table}[!h]
    \centering
    \caption{Performance of combination system with different cutoffs on DUC2002 articles. The last column shows the number (percentage) of correct predicted samples of the combination system. }
    \label{tab:DUC_comp_acc}
    \begin{tabular}{c | c | c | c | c | c || c }
    \hline
    & \multirow{2}{*}{cutoff} & \multirow{2}{*}{\# of samples}
        & \multicolumn{4}{c}{Human Judgement} \\ 
    \cline{4-7}
    &   &  & Icsisumm & Tie & Leadsumm & Combination System \\
    \hline
      & 0.5 & 5  & 4  & 0 & 1  & 246 (78.6\%) \\
      & 0.4 & 13 & 9  & 0 & 4  & 248 (79.2\%) \\
DUC   & 0.3 & 22 & 11 & 1 & 10 & 244 (78.0\%) \\
AP    & 0.2 & 38 & 15 & 1 & 22 & 236 (75.4\%) \\
      & 0.1 & 72 & 21 & 2 & 49 & 215 (68.7\%) \\
      & 0   & 117& 30 & 5 & 82 & 191 (61.0\%) \\
      \cline{2-7}
      & All & 313& 51 (16.3\%) & 19 & 243 (77.6\%) & N/A \\
    \hline\hline
      & 0.5 & 4  & 3  & 0 & 1  & 134 (74.4\%) \\
      & 0.4 & 6  & 5  & 0 & 1  & 135 (75.0\%) \\
DUC   & 0.3 & 8  & 6  & 1 & 1  & 136 (75.6\%) \\
Other & 0.2 & 9  & 6  & 2 & 1  & 136 (75.6\%) \\
      & 0.1 & 18 & 9  & 4 & 5  & 135 (75.0\%) \\
      & 0   & 48 & 19 & 5 & 27 & 123 (68.3\%) \\
      \cline{2-7}
      & All & 180& 41 (22.7\%) & 8 & 131 (72.8\%) & N/A \\
    \hline
    \end{tabular}
\end{table}

Again, we then apply content-dense detector on the generated IcsiSumm and LeadSumm to detect which summary is better based on their content-dense scores.
Table \ref{tab:DUC_comp_acc} shows the results of detecting better IcsiSumm on the DUC2002 articles. 

The judgements on data drawn from sources different from the NYT allows us to get a sense about the extent to which the content-density detector we developed to the news genre in general rather than specifically to the NYT. The observation that bears special mention is that here, the percentage of articles for which the ICSI summarizer is able to produce more informative summaries than the lead summarizer is considerably smaller than in the randomly selected sample of NYT articles. For the AP articles, the automatic summaries are judged as better than the lead baseline for 16\% of the test articles. This conforms with expectations that the AP articles and leads will be overall more content-dense than regular newspaper sources.  For the other sources (newspaper), the percentage of automatic summaries that are better than the lead is 23\%. For comparison, in the NYT sample the system produced a more informative summary in 34\% of the cases. This larger percentage may reflect the style of the New York Times or the fact that the articles from NYT were randomly drawn, so covers a broader range of domains than the DUC data.

The numbers indicate that the style of AP is the most typically informational while the NYT is the most stylistically rich, with leads that are often not content-dense. If this is the case, the room for expediting news search and browsing via automatic summarization has been underestimated in DUC evaluations. 
 
For all three types of sources (AP, NYT, other newspapers), we examine the ability of the content-dense predictor we developed to guide system combination.

As described above, we first ran IcsiSumm and LeadSumm systems on each article,
and then apply the content-density detector on the generated summaries.
Let $Score_{icsi}$ and $Score_{lead}$ denote the content-dense scores for IcsiSumm and LeadSumm respectively,
and $Score_{diff} = Score_{icsi} - Score_{lead}$.
The combination system picks the Icsisumm summary as ouput summary if $Score_{diff} >= cutoff$, otherwise the LeadSumm is picked.
We evaluate the performance for different cutoff thresholds.
The system which always picks the ICSISumm summary and the system always picking lead summary are employed as baselines.

The results of are show in table~\ref{tab:NYT_comp_acc} and~\ref{tab:DUC_comp_acc}.
The first column represents the cutoff value and the second column shows the number of total samples within this cutoff.
Column 3 to 5 are the statistics of human judgements.
The last column shows the number (percentage) of correctly predicted samples of the combination system.
Each row shows the results of the system with a cutoff value. 
The last row shows the statistics for the entire dataset and two baselines,
one that picks ICSISumm summaries only and another that picks lead summaries only.
Not surprisingly, the system always picking lead summary is better than the system picking icsi summary in all three type of sources.
Which is aligned with early studies that lead summary is a hard-to-beat baseline.

The combination system performs better than baseline systems when we pick the right cutoff.
The choice of cutoff depends on the source types and the reason is the writing styles can be very different in different sources as discussed above.
For NYT and DUC other newspapers, the combination system is able the achieve better performance when setting cutoff as 0.1 and achieve highest accuracy when setting cutoff as 0.2.
For the AP, however, it is rather harder to find a cut off in which the combination system would outperfrom the lead baseline.
The cutoff has to be set to 0.4 to get the best performance, so the general ability to produce a better summary with AP data is dubious..


\section{Conclusion and future work}

In this paper we introduced the task of detecting content-dense news article leads. We use article/summary pairs from the NYT corpus to heuristically label a large set of articles as content-dense when the lead of the article overlaps highly with the human summary and as non content-dense when the overlap is low.   

We present experiments with two lexical representations and one syntactic representation. The production rule syntactic representation is the best predictor of lead content-density among the three. The corpus-independent lexical representation from a vocabulary defined by the MRC lexicon proved to be the more useful lexical representation. We compared a feature-level combination model and a two-layer decision-level combination model. The latter performs best in all our experiments. 

Our analysis reveals that there is a large variation across news domains in the fraction of content-dense leads and in the prediction accuracy that can be achieved. Contrary to popular assumptions in news summarization, we find that a large fraction of leads are in fact not content-dense and thus do not provide a satisfactory summary.

Overall domain-specific models are more accurate than in-domain labels from trained annotators. The general model trained on all data pooled together achieves better performance on crowdsourced annotations in both domain-dependent and domain independent annotation conditions. Our experiments indicate that predicting content-dense in terms of real-value scores may be more accurate and beneficial for applications than simply classifying a lead as content-dense or not.

In this work, we have established the feasibility of the task of detecting content-dense texts. We have confirmed that the automatic annotation of data captures distinctions in informativeness as perceived by people. We also show proof-of-concept experiments that show how the approach can be used to improve single-document summarization of news and the generation of summary snippets in news-browsing applications. In future work the task can be extended to more fine-grained levels, with predictions on sentence level and the predictor will be intergared in a fully functioning summarization system.  

All data for the work presented in this paper and the domain-dependent and general classifiers will be made publicly   with the publication of this article. 

\bibliography{informative}
\bibliographystyle{theapa}

\appendix
\section{Reference leads used in AMT annotations}
Here we present all the references leads annotators saw in AMT human intelligent tasks(HITs). 
\subsection{In-domain reference leads}
In in-domain annotations, annotators labeled a group of five leads from same domain in each HIT. Two content-dense leads and two non content-dense leads from the same domain are displayed at the beginning. 

\paragraph{Reference leads for business domain} \mbox{}\\ 
\vspace{-2mm}

\textbf{[Content-dense Ref 1]}
Securities regulators charged one of the richest men in Mexico, Ricardo B. Salinas Pliego, with fraud yesterday, in a lawsuit that seeks to have him barred as a director or officer of any company whose shares trade on an American exchange.

The Securities and Exchange Commission also sought to have Mr. Salinas Pliego, the chairman of TV Azteca, the second-biggest Spanish-language broadcaster, give up more than \$110 million he made from trading in the company's stock and debt.

\vspace{2mm}	\textbf{[Content-dense Ref 2]} 
In a rare move, Microsoft said yesterday that it had agreed to pay a percentage of the sales of its new portable media player to the Universal Music Group.

Universal Music, a unit of Vivendi, will receive a royalty on the Zune player in exchange for licensing its recordings for Microsoft's new digital music service, the companies said. 

\vspace{2mm}	\textbf{[Non content-dense Ref 1]} 
LOOKING for some thong underwear or perhaps a leather jacket and don't know where to find them? Try logging on to a restaurant Web site.

Small restaurateurs are increasingly using the Internet to sell goods that go far beyond the usual array of branded T-shirts and hats, in hopes of not just building the bottom line, but also cultivating possible new markets for expansion.

\vspace{2mm}	\textbf{[Non content-dense Ref 2]} 
''WHAT stresses me most,'' the chief executive of Novartis, Daniel L. Vasella, said, ''is that we are getting new regulations from abroad without any consultation.`` 

This has been the World Economic Forum that the United States government largely passed by. In a world that both respects and fears American power, there is worry that the United States does not care what others think.

\vspace{2mm}
\paragraph{Reference leads for science domain} \mbox{}\\ 
\vspace{-2mm}

\textbf{[Content-dense Ref 1]}
Scientists have decoded the chimp genome and compared it with that of humans, a major step toward defining what makes people human and developing a deep insight into the evolution of human sexual behavior.

The comparison pinpoints the genetic differences that have arisen in the two species since they split from a common ancestor some six million years ago.

\vspace{2mm}	\textbf{[Content-dense Ref 2]}    
A popular class of drugs for high blood pressure, ACE inhibitors, may cause birth defects if taken during the first three months of pregnancy, doctors are reporting. Pregnant women and those who are planning to become pregnant should avoid the drugs, the researchers and officials at the Food and Drug Administration warn.

ACE inhibitors have long been known to cause birth defects if taken later in pregnancy, but until now were considered safe if taken in the first trimester.

\vspace{2mm}	\textbf{[Non content-dense Ref 1]} 
To gauge the potential consumer impact of the consolidation sweeping the telephone industry, look no further than the silver-toned plastic phone gathering dust on the desk in Justin Martikovic's studio apartment.

Mr. Martikovic, 30, a junior architect who relies on a cellphone for his normal calling, says he never uses the desk phone -- but he pays \$360 a year to keep it hooked up.

\vspace{2mm}	\textbf{[Non content-dense Ref 2]}    
As the horror of the South Asian tsunami spread and people gathered online to discuss the disaster on sites known as Web logs, or blogs, those of a political bent naturally turned the discussion to their favorite topics.

To some in the blogosphere, it simply had to be the government's fault.

\vspace{2mm}
\paragraph{Reference leads for sports domain} \mbox{}\\ 
\vspace{-2mm}

\textbf{[Content-dense Ref 1]}
Ivor G. Balding, one of three British brothers who gained international fame as polo stars in the 1930's, when the sport attracted large crowds and wide press coverage, died on Thursday at his home in Camden, S.C. He was 96.

His death was announced by his family.

\vspace{2mm}	\textbf{[Content-dense Ref 2]}    
Finally, the deal is done.
    
Laveranues Coles, the wide receiver from the Washington Redskins, passed a physical examination by the Jets' medical staff yesterday, clearing the way for the team to reacquire him in a trade for wide receiver Santana Moss.

\vspace{2mm}	\textbf{[Non content-dense Ref 1]} 
NEARLY 36 years ago, when it was his turn to interview the prospective employee, the estimable James Reston, onetime traveling secretary for the Cincinnati Reds but then the executive editor of this newspaper, asked how a political science major had wound up writing about sports.

I answered the question, but I have a better answer now. The political science classes prepared me for the nonsense that will pass for a hearing about steroids use in baseball next Thursday in Washington.

\vspace{2mm}	\textbf{[Non content-dense Ref 2]}    
Three years ago, as he stood in the rubble of the St. Bonaventure basketball program, Ahmad Smith had a decision to make.     

One of his teammates, center Jamil Terrell, had been declared ineligible after it was learned that he had been admitted to the Franciscan university in the hills of southwestern New York with a welding certificate -- and the approval of St. Bonaventure's president.

\vspace{2mm}
\paragraph{Reference leads for politics domain} \mbox{}\\ 
\vspace{-2mm}

\textbf{[Content-dense Ref 1]}
 At least 844 American service members were killed in Iraq in 2005, nearly matching 2004's total of 848, according to information released by the United States government and a nonprofit organization that tracks casualties in Iraq.
 
 The deaths of two Americans announced by the United States military on Friday -- a marine killed by gunfire in Falluja and a soldier killed by a roadside bomb in Baghdad -- brought the total killed since the war in Iraq began in March 2003 to 2,178. The total wounded since the war began is 15,955.

\vspace{2mm}	\textbf{[Content-dense Ref 2]}    
Seventeen people died in two separate violent incidents on Sunday and Monday that underscored an increasing sense of lawlessness in Mexico.

A former soldier went on a rampage in a Pacific coast town on Sunday, killing 12 people before local residents chased him down and the police shot him in the town square. Thirteen hours later, gunmen attacked gamblers at an illegal cockfight at a Guadalajara racetrack, killing 4 and wounding 27 when they tossed two grenades into the crowd.

\vspace{2mm}	\textbf{[Non content-dense Ref 1]} 
 President Bush on Tuesday pressed Senate Republican leaders to continue fighting to confirm John R. Bolton as ambassador to the United Nations, even though Senator Bill Frist, the majority leader, said his options had been exhausted and some Republicans urged the appointment of Mr. Bolton when Congress recesses.
 
 ''The president made it very clear that he expects an up-or-down vote,`` Dr. Frist told reporters after meeting with the president. Back in the Capitol, he added, ''I don't want to close that door yet.``

\vspace{2mm}	\textbf{[Non content-dense Ref 2]}    
ARE things getting better or worse in Iraq? That is the basic question, on which much hinges for the United States and the world. Here are some impressionistic answers.

Just over a year ago, on my last visit to the country, I was able to drive north to Tikrit, Saddam Hussein's home town, and south to the Shiite holy city of Najaf. These were not excursions for sitting back and enjoying the scenery. But they were feasible, at high speed and with some risk.

\subsection{Domain-independent annotation}

In  domain-independent annotations, annotators are given a group of five leads randomly selected from all domains. Two informative leads and two uninformative leads are given as references.

\textbf{[Content-dense Ref 1]}
Securities regulators charged one of the richest men in Mexico, Ricardo B. Salinas Pliego, with fraud yesterday, in a lawsuit that seeks to have him barred as a director or officer of any company whose shares trade on an American exchange.

The Securities and Exchange Commission also sought to have Mr. Salinas Pliego, the chairman of TV Azteca, the second-biggest Spanish-language broadcaster, give up more than \$110 million he made from trading in the company's stock and debt.

\vspace{2mm}	\textbf{[Content-dense Ref 2]}    
In a rare move, Microsoft said yesterday that it had agreed to pay a percentage of the sales of its new portable media player to the Universal Music Group.

Universal Music, a unit of Vivendi, will receive a royalty on the Zune player in exchange for licensing its recordings for Microsoft's new digital music service, the companies said.

\vspace{2mm}	\textbf{[Non content-dense Ref 1]} 
 LOOKING for some thong underwear or perhaps a leather jacket and don't know where to find them? Try logging on to a restaurant Web site. 
 
 Small restaurateurs are increasingly using the Internet to sell goods that go far beyond the usual array of branded T-shirts and hats, in hopes of not just building the bottom line, but also cultivating possible new markets for expansion.

\vspace{2mm}	\textbf{[Non content-dense Ref 2]} 
As the horror of the South Asian tsunami spread and people gathered online to discuss the disaster on sites known as Web logs, or blogs, those of a political bent naturally turned the discussion to their favorite topics. 

To some in the blogosphere, it simply had to be the government's fault.

\section{Production Rules with Highest Weights}
\label{sec:appendix_b}
In this section we list the production rules with highest weights for each genre.
We also show two example for each production rule. 
The examples are extracted from lead texts using Stanford CoreNLP package. 

\begin{table}[!htb]
    \centering
    \caption{Top 10 production rules with examples for Business}
    \small
    \label{tab:pr_bu}
    \begin{tabular}{| L{16cm} |}
    \hline
    \multicolumn{1}{|c|}{
    \begin{normalsize}
    \textbf{Positive Rules}
    \end{normalsize}
    } \\ \hline
\textbf{$^{+}$VP-\textgreater VB NP PRT ADVP} \\ \hline
          1) VP -\textgreater VB[scare] NP[them] PRT[away] ADVP[all over again] \\
          2) VP -\textgreater VB[push] NP[the Czech currency] PRT[up] ADVP[sharply] \\ \hline
\textbf{$^{+}$VP-\textgreater VBG PP S} \\ \hline
      1) VP -\textgreater VBG[boasting] PP[on line about their incentive packages] S[to attract companies to relocate to their areas] \\
      2) VP -\textgreater VBG[looking] PP[for facts about different regions] S[to get information that only used to be available , if at all , through the mail and in-person visits] \\ \hline
\textbf{$^{+}$NP-\textgreater NN} \\ \hline
      1) NP -\textgreater NN[response] \\
      2) NP -\textgreater NN[overdrive] \\ \hline
\textbf{$^{+}$VP-\textgreater ADJP VBG NP PP} \\ \hline
      1) VP -\textgreater ADJP[tough] VBG[protecting] NP[American industry] PP[from unfair trading practices] \\
      2) VP -\textgreater ADJP[sometimes heated] VBG[questioning] NP[Tuesday] PP[from members of a House subcommittee] \\ \hline
\textbf{$^{+}$ NP-\textgreater DT NNP} \\ \hline
      1) NP -\textgreater DT[the] NNP[I.M.F.] \\
      2) NP -\textgreater DT[the] NNP[F.D.A.] \\ \hline
      \hline
      \multicolumn{1}{|c|}{
      \begin{normalsize}
      \textbf{Negative Rules}
      \end{normalsize}
      } \\ \hline
\textbf{$^{-}$ NP-\textgreater JJ CD NNS} \\ \hline
      1) NP -\textgreater JJ[pre-April] CD[15] NNS[blues] \\
      2) NP -\textgreater JJ[past] CD[150] NNS[degrees] \\ \hline
\textbf{$^{-}$ VP-\textgreater VBN PP NP-TMP PP} \\ \hline
      1) VP -\textgreater VBN[injured] PP[in a car crash in Peru , a third weathered] NP-TMP[a summer] PP[in Pakistan in brutal 117-degree heat] \\
      2) VP -\textgreater VBN[swayed] PP[down the wet black runway at the Alexander McQueen fashion show last Thursday] NP-TMP[night] PP[to an ominous disco] \\ \hline
\textbf{$^{-}$ ADVP-\textgreater RBR RB PP} \\ \hline
    1) ADVP -\textgreater  RBR[more] RB[often] PP[than not] \\
    2) ADVP -\textgreater  RBR[More] RB[often] PP[than not] \\ \hline
\textbf{$^{-}$ VP-\textgreater VBZ : NP} \\ \hline
    1) VP -\textgreater  VBZ[War] :[:] NP[Has Newsweek 's Time Finally Come] \\
    2) VP -\textgreater  VBZ[is] :[:] NP[Now what] \\ \hline
\textbf{$^{-}$ VP-\textgreater VBD ADVP NP-TMP , NP} \\ \hline
    1) VP -\textgreater  VBD[fell] ADVP[sharply] NP-TMP[yesterday] ,[,] NP[the fourth consecutive decline , as concerns about inflation and interest rates grew before today 's report on producer prices] \\
    2) VP -\textgreater  VBD[opened] ADVP[here] NP-TMP[Friday] ,[,] NP[another sign of how companies all over the world are still rushing to do business in China] \\ \hline
    \end{tabular}
\end{table}

\begin{table}[!htb]
    \centering
    \caption{Top 10 production rules with examples for Science}
    \small
    \label{tab:pr_sc}
    \begin{tabular}{| L{16cm} |}
    \hline
    \multicolumn{1}{|c|}{
    \begin{normalsize}
    \textbf{Positive Rules}
    \end{normalsize}
    } \\ \hline
\textbf{$^{-}$ QP-\textgreater JJR IN NP} \\ \hline
    1) QP -\textgreater  JJR[more] IN[than] NP[the vast majority] \\
    2) QP -\textgreater  JJR[more] IN[than] NP[a jubilant return] \\ \hline
\textbf{$^{-}$ ADJP-\textgreater ADJP SBAR} \\ \hline
    1) ADJP -\textgreater  ADJP[less likely than others to have children , and those who do give birth run an increased risk of bearing a child with the same birth defect] SBAR[that they themselves have] \\
    2) ADJP -\textgreater  ADJP[far less successful] SBAR[than expected] \\ \hline
\textbf{$^{-}$ NP-\textgreater DT NNP NNS NN} \\ \hline
    1) NP -\textgreater  DT[A] NNP[Federal] NNS[appeals] NN[court] \\
    2) NP -\textgreater  DT[a] NNP[Texas] NNS[appeals] NN[court] \\ \hline
\textbf{$^{-}$ S-\textgreater FRAG  NP VP .} \\ \hline
    1) S -\textgreater  FRAG[In] NP[Old] VP[Souls : The Scientific Evidence For Past Lives , '' ( Simon and Schuster , 1999 ) Tom Shroder , a Washington Post editor , reviews the 80-year-old clinical psychiatrist 's research on reincarnation and finds it hard to refute] .[.] \\
    2) S -\textgreater  FRAG[Tonight , when] NP[Live From Lincoln Center ''] VP[broadcasts a concert by the New York Philharmonic on PBS stations across the country , the announcer will not be saying anything about the personal story of the bass-baritone Thomas Quasthoff , who will sing four concert arias by Mozart] .[.] \\ \hline
\textbf{$^{-}$ NP-\textgreater PRP\$ NNS NN} \\ \hline
    1) NP -\textgreater  PRP\$[their] NNS[doctors] NN[charge] \\
    2) NP -\textgreater  PRP\$[their] NNS[employees] NN[home] \\ \hline
    \hline
    \multicolumn{1}{|c|}{
    \begin{normalsize}
    \textbf{Negative Rules}
    \end{normalsize}
    } \\ \hline
\textbf{$^{-}$ VP-\textgreater VBG NP PP PP } \\ \hline
    1) VP -\textgreater  VBG[ordering] NP[a cup of coffee] PP[at Starbucks] PP[into an Olympic challenge] \\
    2) VP -\textgreater  VBG[taking] NP[a crack] PP[at his plays] PP[in the form of faithful revivals or loose interpretations] \\ \hline
\textbf{$^{-}$ VP-\textgreater VB NP ADVP , SBAR } \\ \hline
    1) VP -\textgreater  VB[get] NP[both his legs] ADVP[amputated] ,[,] SBAR[even though they had been perfectly healthy] \\
    2) VP -\textgreater  VB[use] NP[her niece 's card] ADVP[here] ,[,] SBAR[since she does n't live in Westchester] \\ \hline
\textbf{$^{-}$ VP-\textgreater VBN NP , ADVP PP } \\ \hline
    1) VP -\textgreater  VBN[triggered] NP[copycats] ,[,] ADVP[sometimes] PP[by the dozens] \\
    2) VP -\textgreater  VBN[been] NP[7,000 cases of leprosy in this country over the previous three years] ,[,] ADVP[far more than] PP[in the past] \\ \hline
\textbf{$^{-}$ NP-\textgreater NP ,  NP CC NP } \\ \hline
    1) NP -\textgreater  NP[social X-rays] ,[,] NP[those rail-thin women who had attained the exalted status that comes from being married to a Master-of-the-Universe investment banker] CC[or] NP[lawyer] \\
    2) NP -\textgreater  NP[your new book] ,[,] NP[Evolution 's Rainbow : Diversity , Gender and Sexuality in Nature] CC[and] NP[People] \\ \hline
\textbf{$^{-}$ SBAR-\textgreater SBAR , RB SBAR } \\ \hline
    1) SBAR -\textgreater  SBAR[all about whom we could persuade to hire us] ,[,] RB[not] SBAR[whom we would deign to work for] \\
    2) SBAR -\textgreater  SBAR[when they are fine] ,[,] RB[only] SBAR[when they are mucked up or obscure] \\ \hline
    \end{tabular}
\end{table}

\begin{table}[!htb]
    \centering
    \caption{Top 10 production rules with examples for Sports}
    \small
    \label{tab:pr_sp}
    \begin{tabular}{| L{16cm} |}
    \hline
    \multicolumn{1}{|c|}{
    \begin{normalsize}
    \textbf{Positive Rules}
    \end{normalsize}
    } \\ \hline
\textbf{$^{+}$ WHNP-\textgreater WP\$ NN NN } \\ \hline
    1) WHNP -\textgreater  WP\$[whose] NN[baseball] NN[career] \\
    2) WHNP -\textgreater  WP\$[whose] NN[return] NN[date] \\ \hline
\textbf{$^{+}$ VP-\textgreater VBD PRT , S } \\ \hline
    1) VP -\textgreater  VBD[left] PRT[off] ,[,] S[decisively winning the featured Copley Cup race of the 27th annual San Diego Crew Classic yesterday for the second consecutive year] \\
    2) VP -\textgreater  VBD[lashed] PRT[out] ,[,] S[accusing the league of racism] \\ \hline
\textbf{$^{+}$ NP-\textgreater CD JJ JJ NN NN } \\ \hline
    1) NP -\textgreater  CD[one] JJ[infamous] JJ[dining] NN[hall] NN[brawl] \\
    2) NP -\textgreater  CD[seven] JJ[consecutive] JJ[first-round] NN[playoff] NN[series] \\ \hline
\textbf{$^{+}$ VP-\textgreater ADVP VBD NP PP SBAR } \\ \hline
    1) VP -\textgreater  ADVP[out 95 seconds into the first round and Golota] VBD[left] NP[the arena] PP[in an ambulance] SBAR[after he lost consciousness in his locker room after the fight] \\
    2) VP -\textgreater  ADVP[quickly] VBD[switched] NP[him] PP[to second base] SBAR[because Chuck Knoblauch could not throw straight] \\ \hline
\textbf{$^{+}$ ADVP-\textgreater JJ } \\ \hline
    1) ADVP -\textgreater  JJ[next] \\
    2) ADVP -\textgreater  JJ[free] \\ \hline
    \hline
    \multicolumn{1}{|c|}{
    \begin{normalsize}
    \textbf{Negative Rules}
    \end{normalsize}
    } \\ \hline
\textbf{$^{-}$ NP-\textgreater NP , CC NP , PP } \\ \hline
    1) NP -\textgreater  NP[Bob Brenly 's use] ,[,] CC[or] NP[overuse] ,[,] PP[of Curt Schilling] \\
    2) NP -\textgreater  NP[Vt.] ,[,] CC[minus] NP[a number of players still participating in the World Cup] ,[,] PP[including Gretzky , who has been Team Canada 's best player but is also the Rangers ' biggest question mark] \\ \hline
\textbf{$^{-}$ NP-\textgreater DT MD CD NN } \\ \hline
    1) NP -\textgreater  DT[a] MD[May] CD[31] NN[deadline] \\
    2) NP -\textgreater  DT[a] MD[March] CD[4] NN[night] \\ \hline
\textbf{$^{-}$ NP-\textgreater NP JJ NNP NN NN } \\ \hline
    1) NP -\textgreater  NP[Maryland 's] JJ[first] NNP[A.C.C.] NN[tournament] NN[championship] \\
    2) NP -\textgreater  NP[the year 's] JJ[first] NNP[Grand] NN[Slam] NN[tournament] \\ \hline
\textbf{$^{-}$ NP-\textgreater PRP\$ } \\ \hline
    1) NP -\textgreater  PRP\$[his] \\
    2) NP -\textgreater  PRP\$[its] \\ \hline
\textbf{$^{-}$ XS-\textgreater JJ IN } \\ \hline
    1) XS -\textgreater  JJ[much] IN[over] \\
    2) XS -\textgreater  JJ[further] IN[than] \\ \hline
    \end{tabular}
\end{table}

\begin{table}[!htb]
    \centering
    \caption{Top 10 production rules with examples for Politics}
    \small
    \label{tab:pr_us}
    \begin{tabular}{| L{16cm} |}
    \hline
    \multicolumn{1}{|c|}{
    \begin{normalsize}
    \textbf{Positive Rules}
    \end{normalsize}
    } \\ \hline
\textbf{$^{+}$ NP-\textgreater DT NNP : NNP NNP } \\ \hline
    1) NP -\textgreater  DT[the] NNP[Editor] :[:] NNP[Philip] NNP[Gourevitch] \\
    2) NP -\textgreater  DT[the] NNP[Editor] :[:] NNP[Henry] NNP[Siegman] \\ \hline
\textbf{$^{+}$ NP-\textgreater DT VBG NNP NNP } \\ \hline
    1) NP -\textgreater  DT[the] VBG[collapsing] NNP[Soviet] NNP[Union] \\
    2) NP -\textgreater  DT[the] VBG[ruling] NNP[Communist] NNP[Party] \\ \hline
\textbf{$^{+}$ VP-\textgreater VBG NP PRT SBAR } \\ \hline
    1) VP -\textgreater  VBG[propelling] NP[a civic debate] PRT[over] SBAR[whether to change the way Americans experience and ultimately build urban public spaces] \\
    2) VP -\textgreater  VBG[provoking] NP[a debate] PRT[about] SBAR[whether American courts would repeat the kinds of rulings that restricted the civil rights of Japanese-Americans during World War II] \\ \hline
\textbf{$^{+}$ VP-\textgreater VP CC VP S } \\ \hline
    1) VP -\textgreater  VP[are being held in Banco Delta Asia in Macao] CC[and] VP[are] S[to be transferred to a North Korean account at the Bank of China] \\
    2) VP -\textgreater  VP[said the contacts were informal] CC[and] VP[had no bearing on the efforts] S[to help him settle in Panama] \\ \hline
\textbf{$^{+}$ ADVP-\textgreater ADVP CC ADVP } \\ \hline
    1) ADVP -\textgreater  ADVP[at least another week] CC[and] ADVP[perhaps longer] \\
    2) ADVP -\textgreater  ADVP[far enough] CC[and] ADVP[well enough] \\ \hline
    \hline
    \multicolumn{1}{|c|}{
    \begin{normalsize}
    \textbf{Negative Rules}
    \end{normalsize}
    } \\ \hline
\textbf{$^{-}$ ADJP-\textgreater JJ CC RB JJ } \\ \hline
    ADJP -\textgreater  JJ[important] CC[but] RB[relatively] JJ[routine] \\
    ADJP -\textgreater  JJ[tragic] CC[but] RB[not] JJ[surprising] \\ \hline
\textbf{$^{-}$ ADVP-\textgreater DT RP } \\ \hline
    ADVP -\textgreater  DT[all] RP[over] \\
    ADVP -\textgreater  DT[all] RP[around] \\ \hline
\textbf{$^{-}$ NP-\textgreater NP NN PP S } \\ \hline
    NP -\textgreater  NP[Asmat Ali Janbaz 's] NN[explanation] PP[for the American military helicopters] S[flying over this isolated mountain valley last Thursday afternoon] \\
    NP -\textgreater  NP[the Chinese Government 's] NN[use] PP[of military force] S[to suppress the 1989 Tiananmen demonstrations] \\ \hline
\textbf{$^{-}$ SBAR-\textgreater WHADJP S } \\ \hline
    SBAR -\textgreater  WHADJP[exactly what] S[you were doing when you heard that Franklin D. Roosevelt had died , or that John F. Kennedy had been shot , or that Martin Luther King Jr. was dead] \\
    SBAR -\textgreater  WHADJP[How delightful] S[it must be these days to be a member of the Chinese Communist Politburo] \\ \hline
\textbf{$^{-}$ NP-\textgreater VBN NNP NNS } \\ \hline
    NP -\textgreater  VBN[suspected] NNP[Qaeda] NNS[members] \\
    NP -\textgreater  VBN[suspected] NNP[Taliban] NNS[fighters] \\ \hline

    \end{tabular}
\end{table}
    
\end{document}